\documentclass[a4paper,english]{svmult}
\usepackage[utf8]{inputenc}
\usepackage{graphicx}
\usepackage{hyperref}
\usepackage{float, subfig}
\usepackage{listings}
\usepackage{amssymb, bm}
\usepackage{amsmath,amsfonts,amssymb}
\usepackage{lipsum}
\usepackage{algorithmic}
\usepackage{xcolor} 
\usepackage[titlenotnumbered,ruled,vlined,noend,linesnumbered]{algorithm2e}
\usepackage{pdfpages}

\begin{document}
\title*{Model based co-clustering\\of mixed numerical and binary data}

\author{Aichetou Bouchareb, 
        Marc Boullé,
        Fabrice Clérot and 
				Fabrice Rossi
        }

\institute{Aichetou Bouchareb, Marc Boullé and  Fabrice Clérot: \at Orange Labs, 2 Avenue Pierre Marzin 22300 Lannion - France, \email{firstname.lastname@orange.com} 
\and Fabrice Rossi, Aichetou Bouchareb: \at SAMM EA 4534 - University of Paris 1 Panthéon-Sorbonne, 90 rue Tolbiac 75013 Paris - France, \email{firstname.lastname@univ-paris1.fr}}
\maketitle

\abstract{Co-clustering is a data mining technique used to extract the underlying block structure between the rows and columns of a data matrix. Many approaches have been studied and have shown their capacity to extract such structures in continuous, binary or contingency tables. However, very little work has been done to perform co-clustering on mixed type data. In this article, we extend the latent block models based co-clustering to the case of mixed data (continuous and binary variables). We then evaluate the effectiveness of the proposed approach on simulated data and we discuss its advantages and potential limits. 
}

\section{Introduction}
The goal of co-clustering is to jointly perform a clustering of rows and a clustering of columns of a data table. Proposed by \cite{good1965} then by \cite{hartigan1975}, co-clustering is an extension of the standard clustering that extracts the underlying structure in the data in the form of clusters of row and clusters of columns.  
The advantage of this technique, over the standard clustering, lies in the  {\it joint  (simultaneous)} analysis of the rows and columns which enables extracting the maximum of information about the interdependence between the two entities. The utility of co-clustering lies in its capacity to create easily interpretable clusters and its capability to reduce a large data table into a significantly smaller matrix having the same structure as the original data. Performing an analysis on the smaller summary matrix enables the data analyst to indirectly study the original data while significantly reducing  the cost in space and computing time.

Since its introduction, many co-clustering methods have been proposed  (for example, \cite{bock1979, church2000, dhillon2003}). These methods differ mainly in the type of data (continuous, binary or contingency data), in the considered hypotheses, the method of extraction and the expected results (hard clustering, fuzzy clustering, hierarchical clustering, etc.).  One of the renowned approaches is the co-clustering using latent block models which is a mixture model based technique where each cluster of rows  or columns is defined by latent variables to estimate  (\cite{govaert2013}). These models extend the use of Gaussian mixture models and Bernoulli mixture models to the context of co-clustering.

Latent block based co-clustering models have therefore been proposed  and validated for numerical, binary, categorical, and contingency data. Nevertheless, to our knowledge, these models have never been applied to mixed data. Actually, real life data is not always either numerical or categorical and an outright  information extraction method is required to handle mixed type data as well as uni-typed data.  Since the majority of data analysis methods are designed for a particular type of input data, the analyst finds himself$\slash$herself	
forced to go through a phase of data pre-processing to transform the data into a uni-type data (often binary) in order to use an appropriate method. Another option is to separately analyze each part of the data (by type) using an appropriate method, then perform a joint interpretation of the results. However, data pre-processing is very likely to result in a loss of information while independently analyzing different parts of the data, using methods that are based on different models, makes the joint interpretation of the results even harder and sometimes the results are simply incoherent.

Mixture models have been used to analyze mixed data in the context of clustering, by  \cite{damien2015} who propose using a latent variable model according to the Gaussian distribution regardless of the data type (numerical, binary, ordinal, or nominal data). However, the use of these models in co-clustering remains uncommon. In this paper, we propose to extend the co-clustering mixture models, proposed by \cite{govaert2003, govaert2008}, to the case of mixed data (with numerical and binary variables) by adopting the same approach of maximum likelihood estimation as the authors.

The remainder of this paper is organized as follows. In Section \ref{p9_LBM}, we start by defining the latent block models and their use in co-clustering. In Section~\ref{p9_contribution}, we extend these models to mixed data co-clustering. Section~\ref{p9_resultats} presents our experimental results on simulated data. Section~\ref{p9_discussion} provides a discussion of the results.  Finally, conclusions and future work are presented in Section~\ref{p9_conclusion}. 

\section{Latent block model based co-clustering}\label{p9_LBM}
Consider the data table $\mathbf{x}=(x_{ij}, i\in I, j\in J)$  where  $I$ is the set of  $n$ objects and $J$  the set of $d$ variables characterizing the objects, defined by the rows and columns of the matrix $\mathbf{x}$, respectively.  The goal of co-clustering is to find a partition $\mathcal{Z}$ of the rows into $g$ groups and a partition $\mathcal{W}$ of the columns into $m$ groups, describing the permutation of rows and columns that defines groups of rows and groups of columns and forms homogeneous $blocks$ at the intersections of the groups. Supposing the number of row clusters and the number of column clusters to be known, an entry $ x_{ij}$ belongs to the block $B_{kl}=(I_k, J_l)$ if and only if the row $x_{i.}$ belongs to the group $I_k$  of rows, and the column $x_{.j}$ belongs to the group $J_l$ of columns. The partitions of the rows and columns can be represented by the binary matrix $\mathbf{z}$ of row affiliations to the row clusters  and the binary matrix  of column affiliations  $\mathbf{w}$, where $z_{ik}=1$ if and only if $x_{i.} \in I_k$ and $w_{jl} = 1$ if and only if $x_{.j} \in J_l$.

The likelihood of the latent block model LMB is given by:

\begin{equation}\label{p9_densite}
f (\mathbf{x}; \theta) =\sum_{(\mathbf{z},\mathbf{w})\in (\mathcal{Z}\times\mathcal{W})}{p((\mathbf{z}, \mathbf{w});\theta)p(\mathbf{x}|\mathbf{z}, \mathbf{w}
; \theta)},
\end{equation}
where   $\theta$ is the set of unknown model parameters, and $(\mathcal{Z}\times\mathcal{W})$ is the set of all possible partitions $\mathbf{z}$ of $I$ and $\mathbf{w}$ of $J$ that fulfill the following LBM hypotheses:

\begin{enumerate} 
\item the existence of a partition of rows into $g$ clusters $ \{I_1, \ldots, I_g\}$ and a partition of columns into $m$ clusters $\{J_1, \ldots, J_m\}$  such that each entry $x_{ij}$, of the data matrix, is the result of a probability distribution that depends only on its row cluster and its column cluster. These partitions can be represented  by latent variables that can be estimated,

\item the memberships of the row clusters and of the column clusters are independent,

\item knowing the cluster memberships, the observed data units are independent (conditional independence to the couple ($\mathbf{z}$, $\mathbf{w}$)).
\end{enumerate}

Under these hypotheses, \if(0)the likelihood of the model is given by 
\[
f (\mathbf{x}; \theta) =\sum_{(\mathbf{z},\mathbf{w})\in
  \mathcal{Z}\times\mathcal{W}}{\prod_{ik}{\pi_k^{z_{ik}}}\,
  \prod_{jl}{\rho_l^{w_{jl}}}\prod_{ijkl}{\varphi_{kl}(x_{ij};
    \alpha_{kl})^{z_{ik}w_{jl}} }},
\]
and \fi 
the log-likelihood of the data is given by: 
{\small
\[
L(\theta) = \log f (\mathbf{x}; \theta)  =\log\;\left(
  \sum_{(\mathbf{z},\mathbf{w})\in
    \mathcal{Z}\times\mathcal{W}}{\prod_{ik}{\pi_k^{z_{ik}}}\,
    \prod_{jl}{\rho_l^{w_{jl}}}\prod_{ijkl}{\varphi_{kl}(x_{ij};
      \alpha_{kl})^{z_{ik}w_{jl}} }}\;\right),
\]
}
where the sums and products over $i, j, k,$ and $l$ have their limits from $1$ to $n, d,
g, $ and $m$, respectively, $\pi_k$ and $\rho_l$  are the proportions of the $k\textsuperscript{th}$ cluster of rows and the  $l\textsuperscript{th}$ cluster of columns, and $\alpha_{kl}$ is the set of parameters specific to the block $B_{kl}$. The likelihood $\varphi_{kl}$ is that of a Gaussian distribution in the case of numerical data and that of a Bernoulli distribution in the case of binary data.

For an $(n\times d)$ data matrix and a partition into $g\times m$ co-clusters, the sum over $\mathcal{Z}\times\mathcal{W}$ would take at least $g^n\times m^d$ operations (\cite{brault2015b}). Directly computing the log-likelihood is infeasible in a reasonable time, preventing therefore a direct application of EM algorithm,  classically used in mixture models.  Thus,  \cite{govaert2008}  use a variational approximation and a Variational Expectation Maximization algorithm for optimization.

\section{LBM based co-clustering of mixed data}\label{p9_contribution}
The latent block model as defined in Section~\ref{p9_LBM} can only be applied to uni-type data. To extend their use to the case of mixed data, we now consider a mixed type data table $\mathbf{x}=(x_{ij}, i\in I, j\in J=J_c\cup J_d)$ where $I$ is the set of $n$ objects characterized by continuous and binary variable, $J_c$ is the set of $d_c$ continuous variables and $J_d$ the set of $d_d$ binary variables. Our goal is to find a partition of rows into $g$ clusters, a partition of the continuous columns into $m_c$ clusters, and a partition of the binary columns into $m_d$ clusters, denoted $\mathcal{Z}$,  $\mathcal{W}_c$ and $\mathcal{W}_d$
respectively. 

Additionally to the previously mentioned LBM hypotheses, we suppose that the partition of rows, the partition of continuous columns and the partition of the binary columns are independent. These partitions are represented by the binary clustering matrices $\mathbf{z}$, $\mathbf{w}_c$, $\mathbf{w}_d$ and by the fuzzy clustering matrices $s$, $tc$ and $td$, respectively. Furthermore, conditionally on $\mathbf{w}_c$, $\mathbf{w}_d$ and $\mathbf{z}$, the data matrix entries $(x_{ij})_{\{i\in I, j\in J\}}$ are supposed independent and there is a  mean, independent from the model, to distinguish the  continuous columns from the binary ones. Under these hypotheses, the likelihood of the generative model for mixed data can be written as: 

{\small
\begin{equation*}
\begin{aligned}
f(\mathbf{x}; \theta) =& \sum_{(\mathbf{z},\mathbf{w}_c, \mathbf{w}_d)\in (\mathcal{Z}\times(\mathcal{W}_c,\mathcal{W}_d))}\left(\prod_{ik}{\pi_k^{z_{ik}}}\, \prod_{j_cl_c}\rho_{{l_c}}^{wc_{j_cl_c}}\prod_{j_dl_d}{\rho_{{l_d}}^{wd_{j_dl_d}}}\right.\\
& {\quad\left.\prod_{ij_ckl_c}{\varphi_{kl_c}^c(x_{ij_c}; \alpha_{kl_c})^{z_{ik}wc_{j_cl_c}} }\prod_{ij_dkl_d}{\varphi_{kl_d}^d(x_{ij_d}; \alpha_{kl_d})^{z_{ik}wd_{j_dl_d}} }\right)}.
\end{aligned}
\end{equation*}}
Note that the aforementioned hypotheses lead to a simple combination of the previously existing situations (binary and continuous). Therefore, this combination  adds no further mathematical difficulty, but rather potential practical consequences, resulting from coupling two different distributions (in the clustering of rows) and by the incommensurable natures of the densities (continuous variables) and probabilities (binary variables).

For likelihood optimization, we use an iterative Variational Expectation Maximization algorithm, inspired by \cite{govaert2008}, as described below. 

\subsection{Variational approximation}

In the latent block model, the goal is to optimize the full-information, which requires knowing the latent variables $\mathbf{z}$, $\mathbf{w}_c$ and $\mathbf{w}_d$. The full-information log-likelihood is given by: 

{\small
\begin{equation*}
\begin{aligned}
&L_c(\mathbf{x}, \mathbf{z}, \mathbf{w}_c, \mathbf{w}_d; \theta )= \sum_{ik}{z_{ik}\log \pi_k}\,+\, \sum_{j_cl_c}{wc_{j_cl_c} \log \rho_{l_c}}\,+\, \sum_{j_dl_d}{wd_{j_dl_d} \log \rho_{l_d}}\,\\
& +\, \sum_{ij_ckl_c}{z_{ik}wc_{j_cl_c} \log \varphi_{kl_c}^c(x_{ij_c}; \alpha_{kl_c})}\,+\, \sum_{ij_dkl_d}{z_{ik}wd_{j_dl_d} \log \varphi_{kl_d}^d(x_{ij_d}; \alpha_{kl_d})},
\end{aligned}
\end{equation*}}
where the sums over $i, j_c, j_d, k, l_c, l_d $ have their limits from $1$ to $n,
d_c, d_d, g, m_c$ and $m_d$ respectively.

However, a direct application of the EM algorithm is impractical due to the dependency between the memberships to the clusters of rows $\mathbf{z}$ and  the memberships to the clusters of continuous columns $\mathbf{w}_c$ on one hand, and between the memberships to the clusters of rows $\mathbf{z}$ and the memberships to the clusters of binary columns 
$\mathbf{w}_d$, on the other hand. This makes the computation of the joint distribution $p(\mathbf{z}, \mathbf{w}_c, \mathbf{w}_d|\mathbf{x}, \theta)$ rather an impossible task. It is thus impractical to integrate the log-likelihood of the full-information data, given this distribution.

As in \cite{govaert2008}, we use a variational approximation that consists of approximating the conditional distributions  of the latent variables to a factorisable form. More precisely, we approximate 
$p(\mathbf{z}, \mathbf{w}_c, \mathbf{w}_d|\mathbf{x}, \theta)$ by the adjustable distribution  product $q(\mathbf{z}|\mathbf{x}, \theta)$,
$q(\mathbf{w}_c|\mathbf{x}, \theta)$ and $q(\mathbf{w}_d|\mathbf{x},
\theta)$, of parameters $s_{ik}=q(z_{ik}=1|\mathbf{x}, \theta$), $tc_{jl}=q(wc_{jl}=1|\mathbf{x}, \theta)$ and $td_{jl}=q(wd_{jl}=1|\mathbf{x}, \theta)$ respectively. 

The full-information likelihood is thus lower bounded by the following $F_c$ criterion 
{\small
 \begin{equation*}
\begin{aligned}
F_c(s, tc, td, \theta)& = \sum_{ik}{s_{ik}\log \pi_k}\,+\, \sum_{j_cl_c}{tc_{j_cl_c} \log \rho_{l_c}}\,+\, \sum_{j_dl_d}{td_{j_dl_d} \log \rho_{l_d}}\, \\
&+\, \sum_{ij_ckl_c}{s_{ik}tc_{j_cl_c} \log \varphi_{kl_c}^c(x_{ij_c}; \alpha_{kl_c})}\,+\, \sum_{ij_dkl_d}{s_{ik}td_{j_dl_d} \log \varphi_{kl_d}^d(x_{ij_d}; \alpha_{kl_d})}\\
& - \sum_{ik}{s_{ik} \log s_{ik}} - \sum_{j_cl_c}{tc_{j_cl_c} \log tc_{j_cl_c}} - \sum_{j_dl_d}{td_{j_dl_d} \log td_{j_dl_d}},
\end{aligned}
\end{equation*}
}

which  provides an approximation for the likelihood. The maximization of $F_c$ is simpler to conduct and yields a maximization of the expected full-information log-likelihood. Therefore, the goal  will onward be to maximize the criterion $F_c$.

\subsubsection{The Variational Expectation Maximization algorithm}
Maximizing the lower bound $F_c$ in the mixed-data latent block model (MLBM) is performed, until convergence, in three steps:  
 
\begin{itemize}
\item[$\bullet$] with regard to $s$, with fixed $\theta$, $tc$ and $td$, which amounts  to computing  
{\small
\begin{equation} \label{p9_S}
 \hat{s}_{ik} \propto \pi_k\exp{(\sum_ {j_cl_c}tc_{j_cl_c}\log \varphi^c_{kl_c}(x_{ij_c}, \alpha_{kl_c}))}\;\exp{(\sum_ {j_dl_d}td_{j_dl_d}\log \varphi^d_{kl_d}(x_{ij_d}, \alpha_{kl_d}))},
\end{equation} 
}
\item[$\bullet$] with regard to $tc$ and $td$ with fixed  $s$ and $\theta$, which amounts to computing  
\begin{equation} \label{p9_TcTd}
\begin{aligned}
&\hat{tc}_{j_cl_c} \propto  \rho_{l_c}\exp{(\sum_ {ik}s_{ik}\log \varphi_{kl_c}^c(x_{ij_c}, \alpha_{kl_c}))} \\
\text{ and } \; &\hat{td}_{j_dl_d} \propto \rho_{l_d}\exp{(\sum_ {ik}s_{ik}\log \varphi_{kl_d}^d(x_{ij_d}, \alpha_{kl_d}))},
\end{aligned}
\end{equation}
with: $\sum\limits_k{s_{ik}} = \sum\limits_{l_c}tc_{jl_c}  = \sum\limits_{l_d}td_{jl_d} =1$,
\item with regard to  $\theta$,  which amounts to computing the cluster proportions and parameters  
\small{
\begin{equation} \label{p9_theta}
\begin{aligned}
 \hat{\pi}_k  & =  \frac{\sum_{i}{\hat{s}_{ik}}}{n}\,; \; \hat{\rho}_{l_c}=\frac{\sum_{j_c}{\hat{tc}_{jl_c}}}{d_c};\; \hat{\rho}_{l_d}=\frac{\sum_{j_d}{\hat{td}_{jl_d}}}{d_d};\; 
\hat{\mu}_{kl_c}=\frac{\sum_{ij_c}{\hat{s}_{ik}\hat{tc}_{j_cl_c}x_{ij_c}}}{\sum_i{\hat{s}_{ik}}\sum_{j_c}{\hat{tc}_{j_cl_c}}}; \\
& \hat{\sigma}_{kl_c}^2=\frac{\sum_{ij_c}{\hat{s}_{ik}\hat{tc}_{j_cl_c}(x_{ij_c}-\hat{\mu}_{kl_c})^2}}{\sum_i{\hat{s}_{ik}}\sum_{j_c}{\hat{tc}_{j_cl_c}}}\, \text{ et }\,\hat{\alpha}_{kl_d}=\frac{\sum_{ij_d}{\hat{s}_{ik}\hat{td}_{j_dl_d}x_{ij_d}}}{\sum_i{\hat{s}_{ik}}\sum_{j_d}{\hat{td}_{j_dl_d}}}.
\end{aligned}
\end{equation} }
\end{itemize}
In our implementation (Algorithm~\ref{p9_LBVEM}),  we used $\epsilon = 10^{-5}$ as convergence constant for the inner  loops, $\epsilon = 10^{-10}$ for the outer loop, and we normalized $\hat{s}$, $\hat{tc}$ and $\hat{td}$, after calculation, by taking the relative values: $\hat{s}_{ik} \leftarrow \frac{\hat{s}_{ik}}{\sum_{h}{\hat{s}_{ih}}}$ and similarly for $\hat{tc}$ and $\hat{td}$.
\begin{algorithm}[ht]
 \caption{The Mixed-data Latent Block Model VEM algorithm}
  \label{p9_LBVEM}
\begin{algorithmic}
\REQUIRE {Data $\mathbf{x}$, the number of clusters $g$, $m_c$, $m_d$, maximum number of iterations $maxITER$ and $InnerMaxIter$} 
\STATE iteration c$\leftarrow$0 
\STATE Initialization: choose $s=c^{c}, \, td=tc^{c}, td=td^{c}  $ randomly and compute   $\theta=\theta^{c}$ (equation \eqref{p9_theta})
\WHILE {$c \le maxITER$ \AND $\;$Unstable($Criterion$)}
	\STATE  t$\leftarrow$0,  $s^t\leftarrow s^c, tc\leftarrow tc^c,td\leftarrow td^c, \theta^t\leftarrow \theta^c$
	 \WHILE {$t \le InnerMaxIter $ \AND Unstable($Criterion$)}
		\STATE \hspace{\algorithmicindent} For every  $i=1:n$ and $ k=1:g$, compute $s_{ik}^{t+1}$:  equation  \eqref{p9_S}
		\STATE \hspace{\algorithmicindent} For every $ k=1:g$, $l_c=1:m_c$ and $l_d=1:m_d$, compute $\pi_k^{t+1}, \mu_{kl_c}^{t+1}, \sigma_{kl_c}^{t+1} \text{ et } \alpha_{kl_d}^{t+1}$:  equation  \eqref{p9_theta}

		\STATE \hspace{\algorithmicindent}$Criterion\leftarrow F_c(s^{t+1}, \,tc, \,td, \,\theta^{t+1})$
		\STATE \hspace{\algorithmicindent}$t\leftarrow t+1$
	\ENDWHILE
		\STATE   $s \leftarrow s^{c+1}\leftarrow s^{t-1},\, \theta \leftarrow \theta^{c+1} \leftarrow\theta^{t-1}$
		\STATE $ t \leftarrow 0 $
	\WHILE {$t \le InnerMaxIter $ \AND Unstable($Criterion$)}
		\STATE \hspace{\algorithmicindent} For every $j_c=1:d_c$, $j_d=1:d_d$, $ l_c=1:m_c$ and $ l_d=1:m_d$, compute $tc_{j_cl_c}^{t+1}$ and $td_{j_dl_d}^{t+1}$:  equation  \eqref{p9_TcTd}
		\STATE \hspace{\algorithmicindent} For every $ k=1:g$, $l_c=1:m_c$ and $l_d=1:m_d$, compute $\rho_c^{t+1},\, \rho_d^{t+1},\, \mu_{kl_c}^{t+1},\, \sigma_{kl_c}^{t+1}$ and $ \alpha_{kl_d}^{t+1}$:  equation  \eqref{p9_theta}

		\STATE \hspace{\algorithmicindent}$Criterion\leftarrow F_c(s,\, tc^{t+1}, \,td^{t+1}, \,\theta^{t+1})$
		\STATE \hspace{\algorithmicindent}$t\leftarrow t+1$
	\ENDWHILE
\STATE {$tc\leftarrow tc^{c+1}\leftarrow tc^{t-1},\, td\leftarrow td^{c+1}\leftarrow td^{t-1},\, \theta \leftarrow \theta^{c+1} \leftarrow\theta^{t-1}$ }
		\STATE $Criterion\leftarrow F_c(s,\, tc,\, td,\, \theta)$

\STATE $c\leftarrow c+1$
\ENDWHILE
\ENSURE $(s, \,tc,\, td, \,\theta)$

 \end{algorithmic}
\end{algorithm}
\section{Experiments}
\label{p9_resultats}
In this section, we evaluate the proposed approach on simulated data with controlled setups. This evaluation step is necessary to measure how well the approach can uncover the true distributions from data with known parameters. To do this, we start by presenting the setups used to produce artificial data followed by an analysis of the  experimental results of the proposed LBM extension. The first experiment is set to validate our implementation on uni-type data and confirm the contribution of the approach. The second experiment is set to investigate the influence of various parameters such as the number of co-clusters, the size of the data matrix and the level of overlap in the data.

\subsection{First experiment}
\label{p9_expe1}
The purpose of this experiment is two-fold: validate our implementation and evaluate the interest of considering continuous and binary data jointly.
\subsubsection{The data set}
Our first data sets consist of  simulated data matrices containing $g=4$ clusters of rows, $m_c=2$ clusters of continuous columns and $m_d=2$ clusters of binary columns. 

The particularity of this experiment lies in the fact that independently co-clustering the continuous and the binary parts of the data would only distinguish two clusters of rows but jointly, the co-clustering of the data sets should extract four clusters of rows.  In this experiment, we study the effect of the size of the data matrix and the level of overlap.
\begin{itemize}
\item {\bf{The size of the data matrix}}: the data size is defined by the number of rows which is  equal to the number of continuous columns and to the number of binary columns. We consider the sizes $25$, $50$, $100$, $200$ and $400$ rows (and columns of each type) for which the resulting matrices will have  $25\times 50$, $50\times 100$, $100\times 200$, $200\times 400$ and  $400\times 800$ entries respectively.

\item {\bf{The level of confusion}} where we study the effect of the overlap between the distributions. Here, we consider three levels of overlap (called confusion) between the co-clusters:
\begin{itemize}
\item  {\it Low}: every continuous co-cluster follows a Gaussian distribution of mean $\mu \in \{\mu_1=1,\, \mu_2=2\}$ and a standard deviation  $\sigma= 0.25$ while a binary co-cluster follows a Bernoulli distribution of parameter $\alpha \in\{\alpha_1=0.2,\, \alpha_2=0.8\}$).  This setup provides easily separable co-clusters since the regions of overlap between the observed values is small. 

\item {\it  Medium}: every continuous co-cluster follows a Gaussian distribution of mean $\mu \in \{\mu_1=1,\, \mu_2=2\}$ and a standard deviation  $\sigma= 0.5$ while a binary co-cluster follows a Bernoulli distribution of parameter $\alpha \in\{\alpha_1=0.3,\, \alpha_2 = 0.7\}$. This setup provides a relatively large overlap region which should make the cluster separability harder than in the case of low confusion.  

\item {\it High}: every continuous co-cluster follows a Gaussian distribution of mean $\mu \in \{\mu_1=1,\, \mu_2=2\}$ and a standard deviation  $\sigma= 1$ while a binary co-cluster follows a Bernoulli distribution of parameter $\alpha \in\{\alpha_1=0.4, \,\alpha_2=0.6\}$.  This provides a large overlap region which should make the cluster separability even more difficult. 

\end{itemize}

\end{itemize}

The exact configuration of the parameters is shown in Table~\ref{p9_configurations1}. One should note that a Gaussian mixture based co-clustering on the  columns $Jc_1$ and $Jc_2$  from Table~\ref{p9_configurations1}, would distinguish two clusters of rows by coupling $\{I_1$ and $I_3\}$, on one hand,  then $\{I_2$ and $I_4\}$,  on the other hand, as single row clusters. Similarly, a Bernoulli based co-clustering on the columns $Jd_1$ and $Jd_2$ should distinguish two clusters  of rows by associating $\{I_1$ with $I_2\}$ and $\{I_3$ with $I_4\}$.  By performing a co-clustering on the mixed data, we expect to distinguish four clusters of rows.

\begin{table}[h]
\begin{center}
\scalebox{0.8}{
\begin{tabular}{cc} 
\begin{tabular}{|l|l|l|l|l|}
  \hline
$\mu$ and  $\alpha$  & $Jc_1$ &$Jc_2$ &$Jd_1$&$Jd_2$\\
  \hline
 $I_1$ & $\mu_2$&  $\mu_1$ &  $\alpha_2$& $\alpha_1$\\
  \hline
  $I_2$ & $\mu_2$ &  $\mu_2$ &  $\alpha_2$& $\alpha_1$\\
\hline
$I_3$ &  $\mu_2$ &  $\mu_1$ &  $\alpha_2$ & $\alpha_2$\\
      \hline
$I_4$ &  $\mu_2$ &  $\mu_2$ &  $\alpha_2$ & $\alpha_2$\\
  \hline
\end{tabular} 
\end{tabular}
}\caption{\small{{\it The specification of the true parameters.}}}\label{p9_configurations1}
 \end{center}
\end{table}

Our experiments are performed in two steps: apply the co-clustering algorithm to the continuous data and to the binary data separately, then apply the algorithm on the mixed data. 

\subsubsection{Evaluation of the results}
Knowing the true clusters of each row and column of the data, we choose to measure the performance of a co-clustering using the Adjusted Rand's Index (\cite{Hubert1985}) for the rows and columns. The Adjusted Rand index (ARI) is a  commonly used measure of similarity between two data clusterings that can be used to measure  the distance (as a probability of  agreements) between the true row and column partitions and the partitions found by the co-clustering. The ARI has a maximum value of 1 for identical partitions and a minimal value of zero for independent partitions.  We will thus  recover  and compare the ARI of rows and columns in the three cases:  when co-clustering the continuous data alone, when co-clustering the binary data alone, and when co-clustering the mixed  data. 

For each configuration, we generate 3 data samples according to the previously illustrated parameters and we present the results in the form of violin plots. A violin plot  (\cite{violin_plot}) is a  numeric data visualization method that 
combines the advantages of a box plots  with  an estimation of the probability density over the different values, which gives a better visualization of the variability of the results as well as important statistics such as the mean, the median and the extent of the measured values.

\subsubsection{Validating our implementation}
\label{p9_validation}
To validate our implementation, we applied our implementation of the co-clustering algorithm to the continuous part alone and to the binary part alone while comparing the results with those of the blockcluster package (\cite{bhatia2014}). Blockcluster is an R package for co-clustering binary, contingency, continuous and categorical data that implements the standard latent block models for co-clustering uni-type data.

\begin{figure}[h]
\centering
\subfloat[ARI of rows, Low confusion]{\includegraphics[ width=0.31\textwidth]{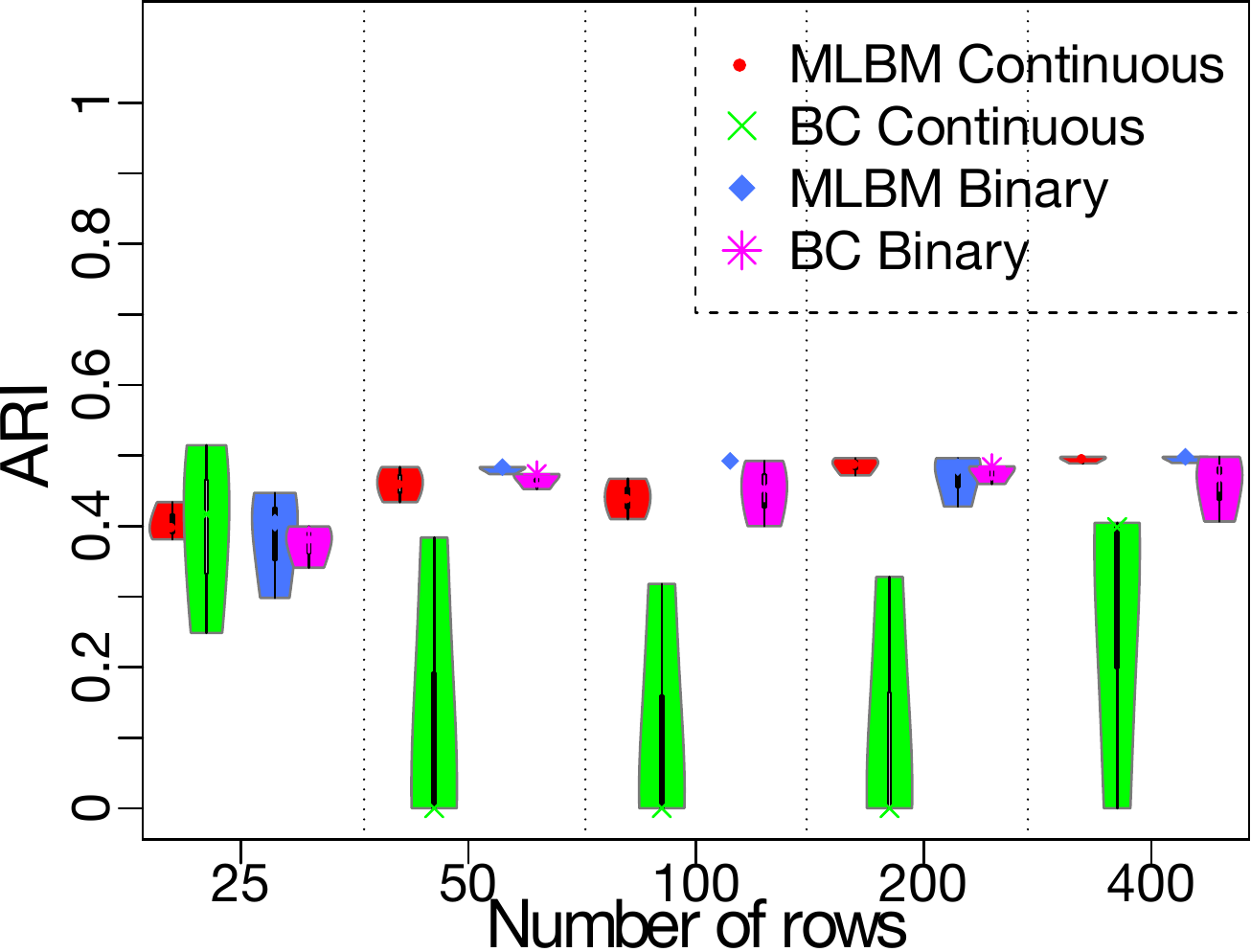}\label{p9_CSGlowR}}
\quad 
\subfloat[ARI of rows, Medium confusion]{\includegraphics[width=0.31\textwidth]{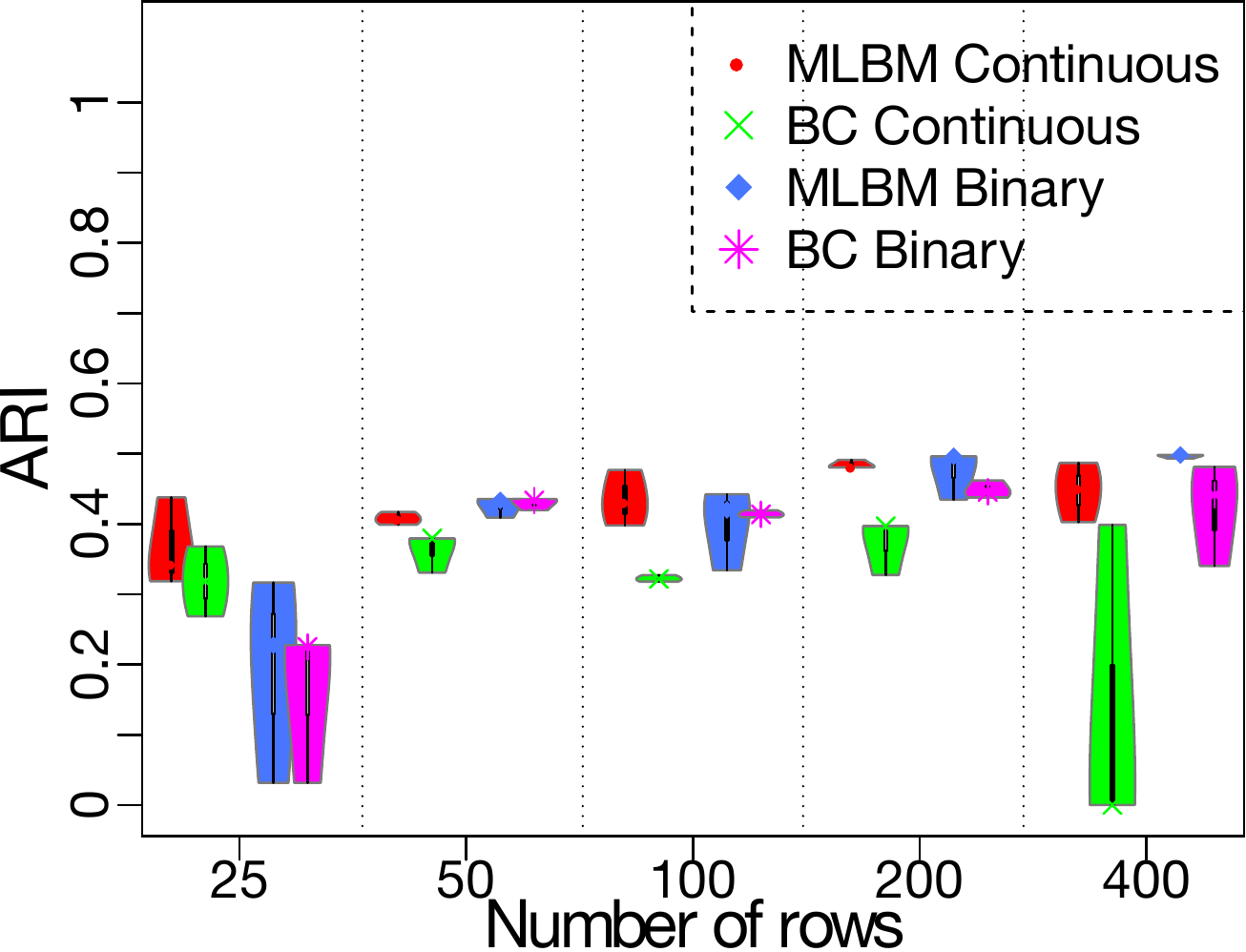}\label{p9_CSGmediumR}}
\quad
\subfloat[ARI of rows, High confusion]{\includegraphics[width=0.31\textwidth]{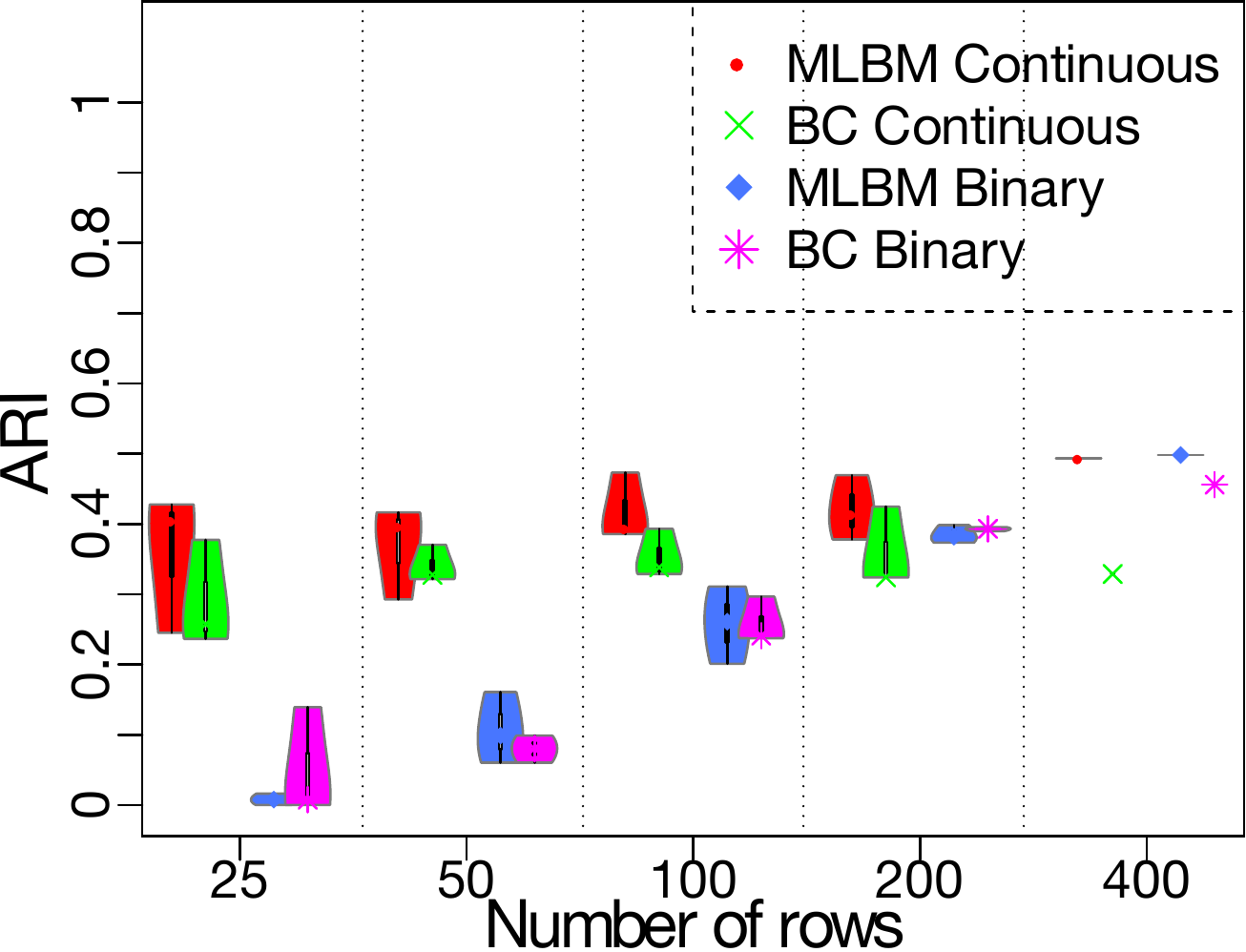}\label{p9_CSGhighR}}
\quad 
\subfloat[ARI of columns, Low confusion]{\includegraphics[ width=0.31\textwidth]{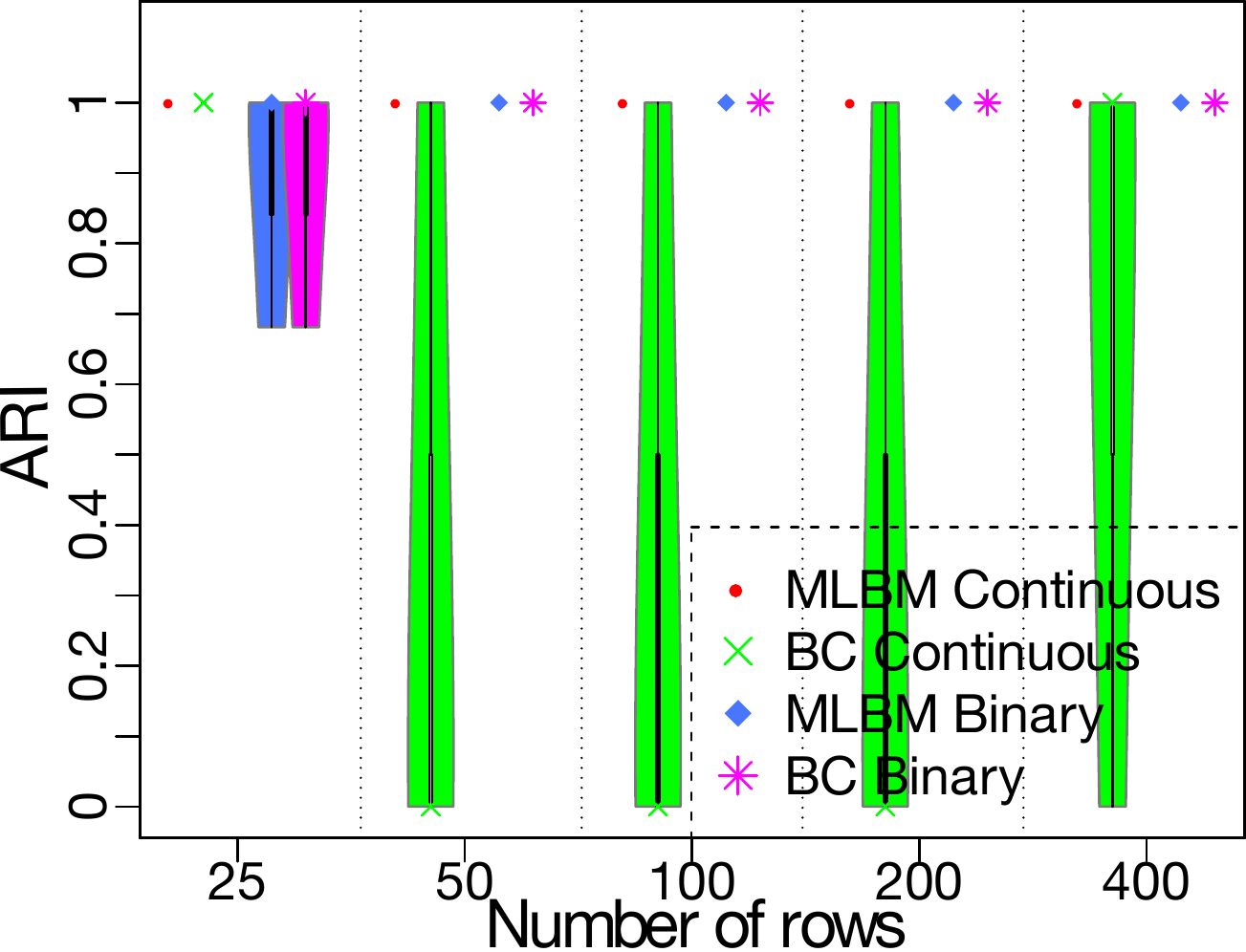}\label{p9_CSGlowC}}
\quad 
\subfloat[ARI of columns, Medium confusion]{\includegraphics[width=0.31\textwidth]{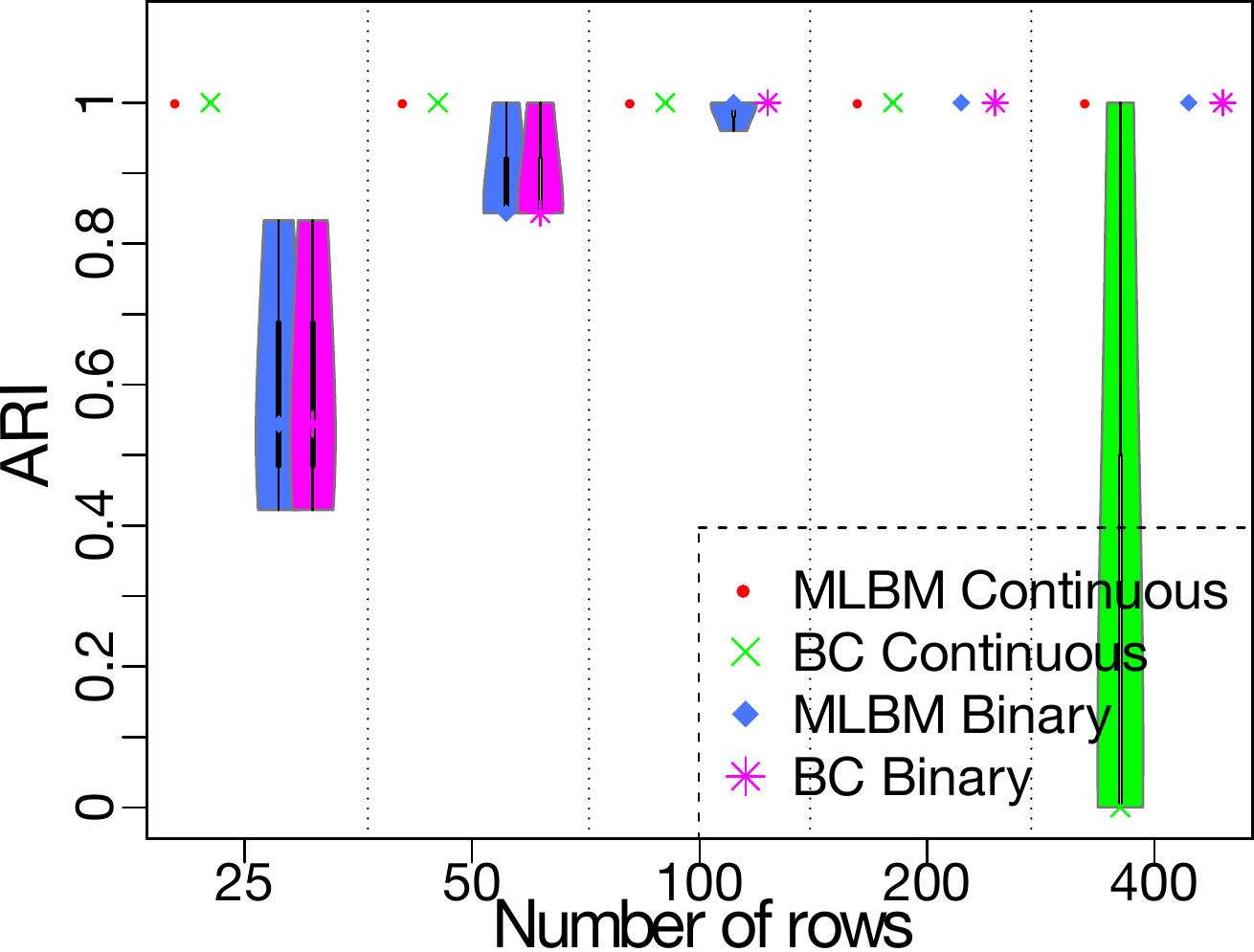}\label{p9_CSGmediumC}}
\quad
\subfloat[ARI of columns, High confusion]{\includegraphics[width=0.31\textwidth]{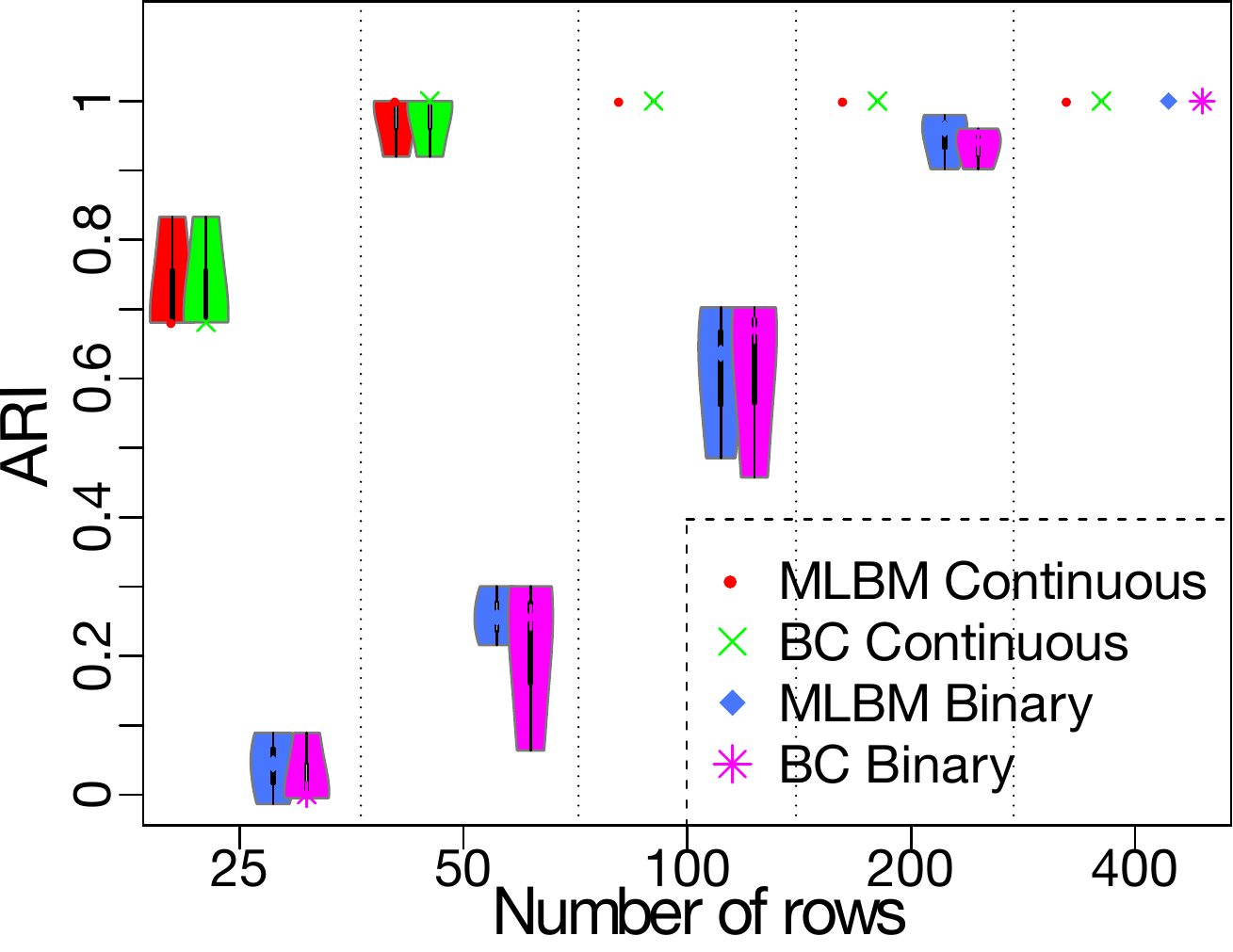}\label{p9_CSGhighC}}
\quad 
\caption{{\it {\small First experiment: comparing the ARI of rows and the ARI of columns (the y-axis) using our implementation (MLBM) with the blockcluster (BC) package, applied to the continuous and binary data. Compare the red plots with the green ones and the blue with the magenta. The higher the ARI values, the better.}}}\label{p9_experiences1SG}
\end{figure}

Figure~\ref{p9_experiences1SG} shows a comparison between the adjusted Rand index (of rows and columns) of the co-clustering obtained using blockcluster compared to our proposed approach. 
 The comparison confirms that our implementation provides  very comparable results, in terms of ARI and of parameter estimation, with respect to the blockcluster package in most of the cases. In particular, BC provides better ARI when co-clustering the binary data while our approach provides similar or remarkably better results when co-clustering the continuous data.  However, in terms of computation time, our implementation takes at least ten times longer than
the blockcluster package. This is mainly because we needed high quality in our comparison experiments, and therefore we focused on quality rather than
computation time in our implementation (see Section~\ref{p9_discussion}).

\subsubsection{The advantage of mixed data co-clustering}\label{p9_advantages}

One approach to co-clustering mixed data consists of performing a co-clustering on each data type then jointly analyzing the results to conclude a co-clustering like structure for the complete data.  This experiment provides an example of configurations where such joint analysis remains incapable of finding the true clusters of rows.

\begin{figure}[htbp]
\centering
\subfloat[Low confusion]{\includegraphics[width=0.31\textwidth]{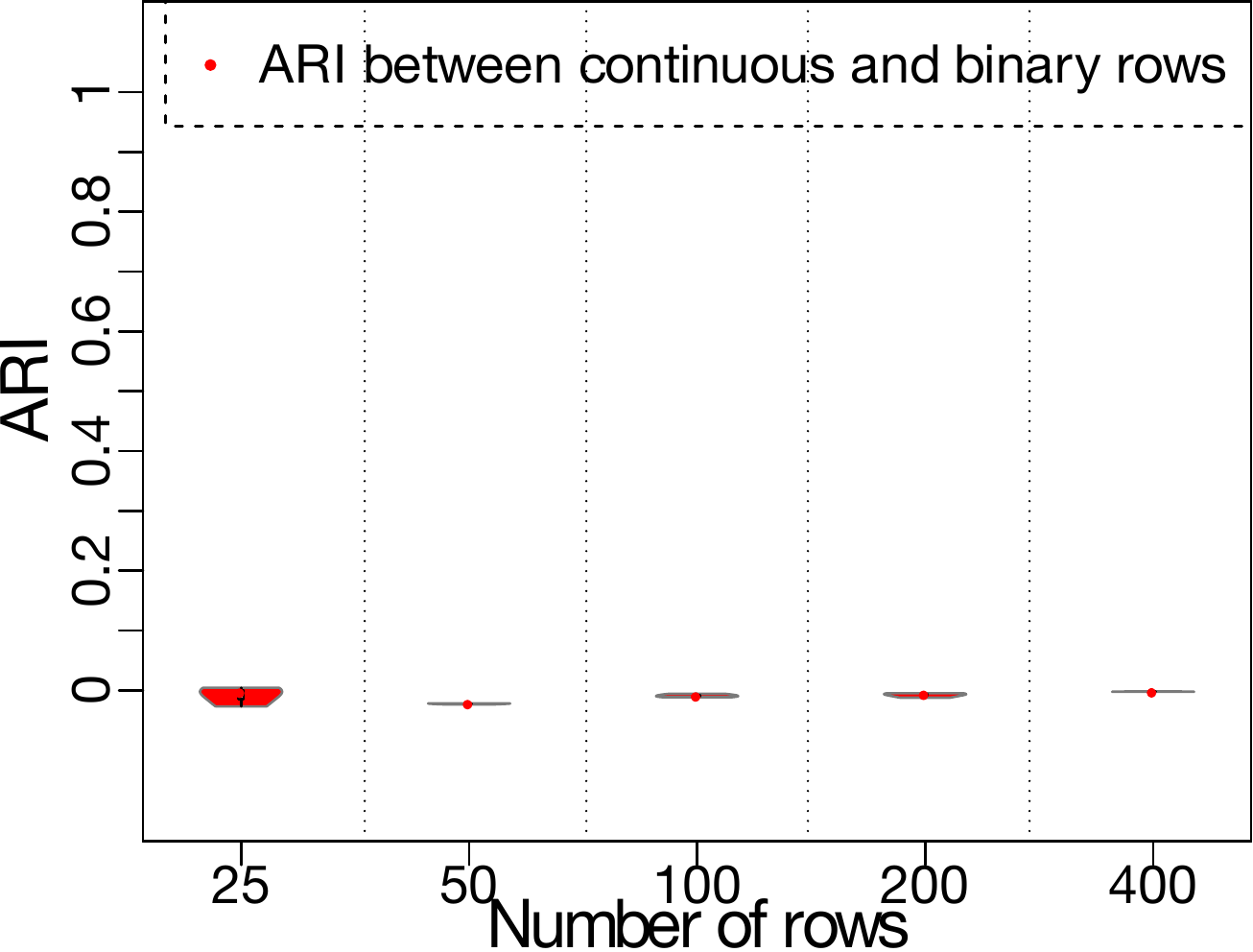}\label{p9_Joint_experiences1lowR}}
\quad
\subfloat[Medium confusion]{\includegraphics[width=0.31\textwidth]{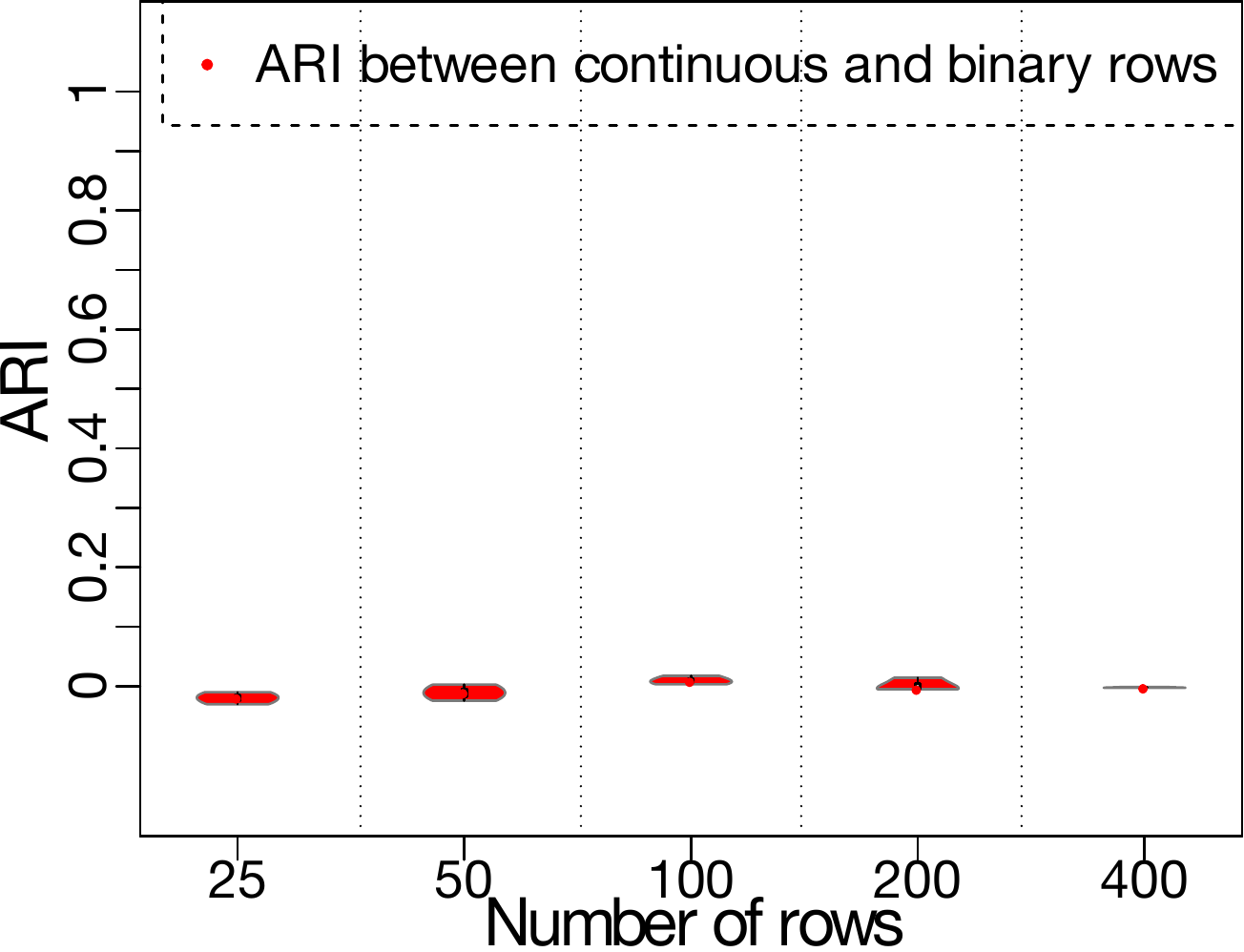}\label{p9_Joint_experiences1mediumR}}
\quad
\subfloat[High confusion]{\includegraphics[width=0.31\textwidth]{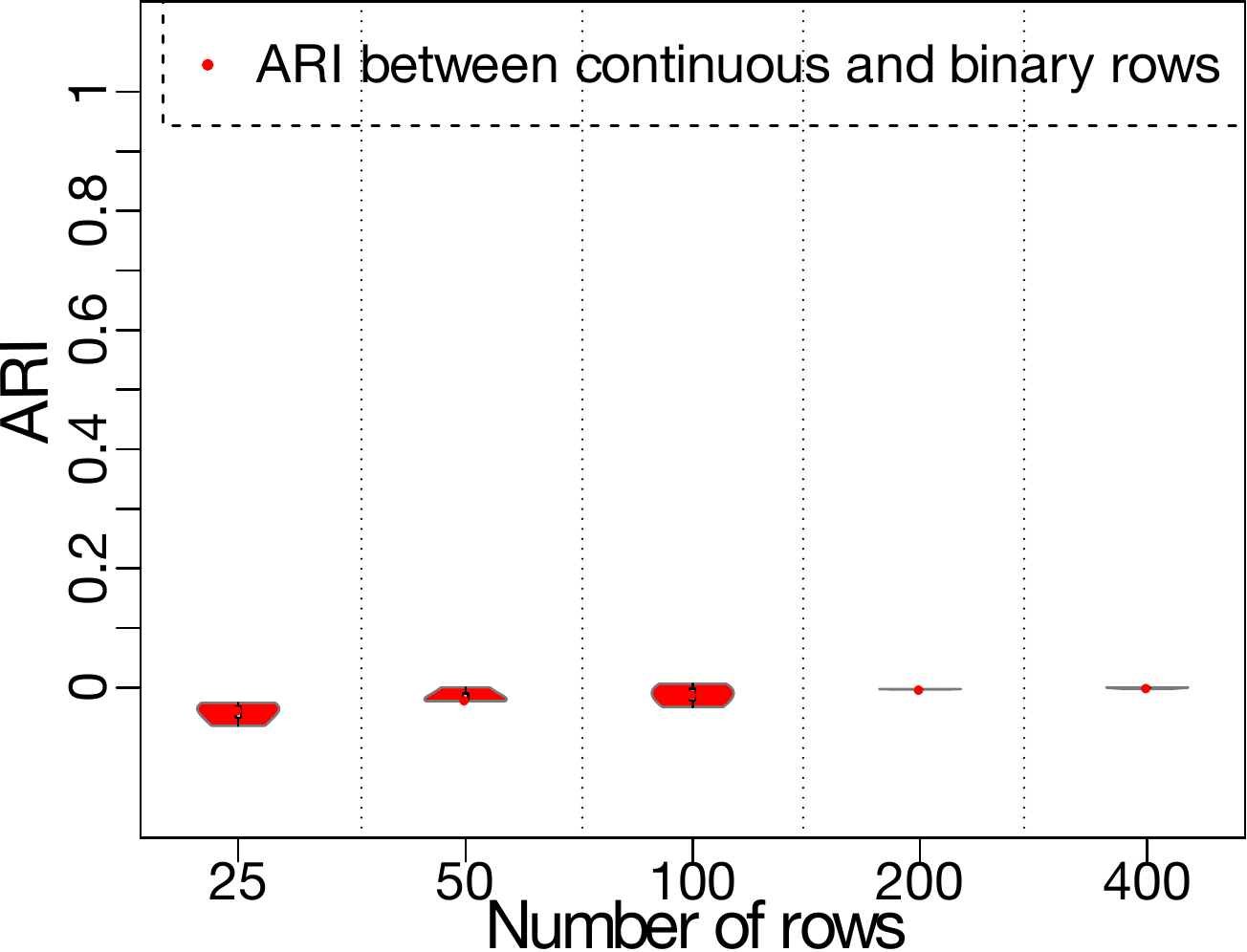}\label{p9_Joint_experiences1highR}}
\caption{{\it {\small First experiment: comparing the partition of rows obtained using the continuous part alone with the partition obtained using the binary part of the data. The y axis shows the measured ARI values.}}}\label{p9_Joint_experiences1}
\end{figure}

Figure~\ref{p9_Joint_experiences1} compares the partition of rows found by the co-clustering of the continuous data with the partition found by the co-clustering of the binary part. Had the two co-clusterings correctly discovered the true clusters of rows, the partitions would be coherent and the  ARI  would approach 1, which is not the case. In fact, regardless of the data size and of the level of overlap between the distributions, the two partitions are completely independent as shown by the ARI values, which are at maximum zero.  This shows that although the same row clusters are present in both data types, the joint analysis of the two independent co-clusterings does not extract the common and global structure and does not provide any additional information on the true distribution compared to a uni-type data analysis. 

\begin{figure}[htbp]
\centering
\subfloat[ARI of rows, Low confusion]{\includegraphics[ width=0.31\textwidth]{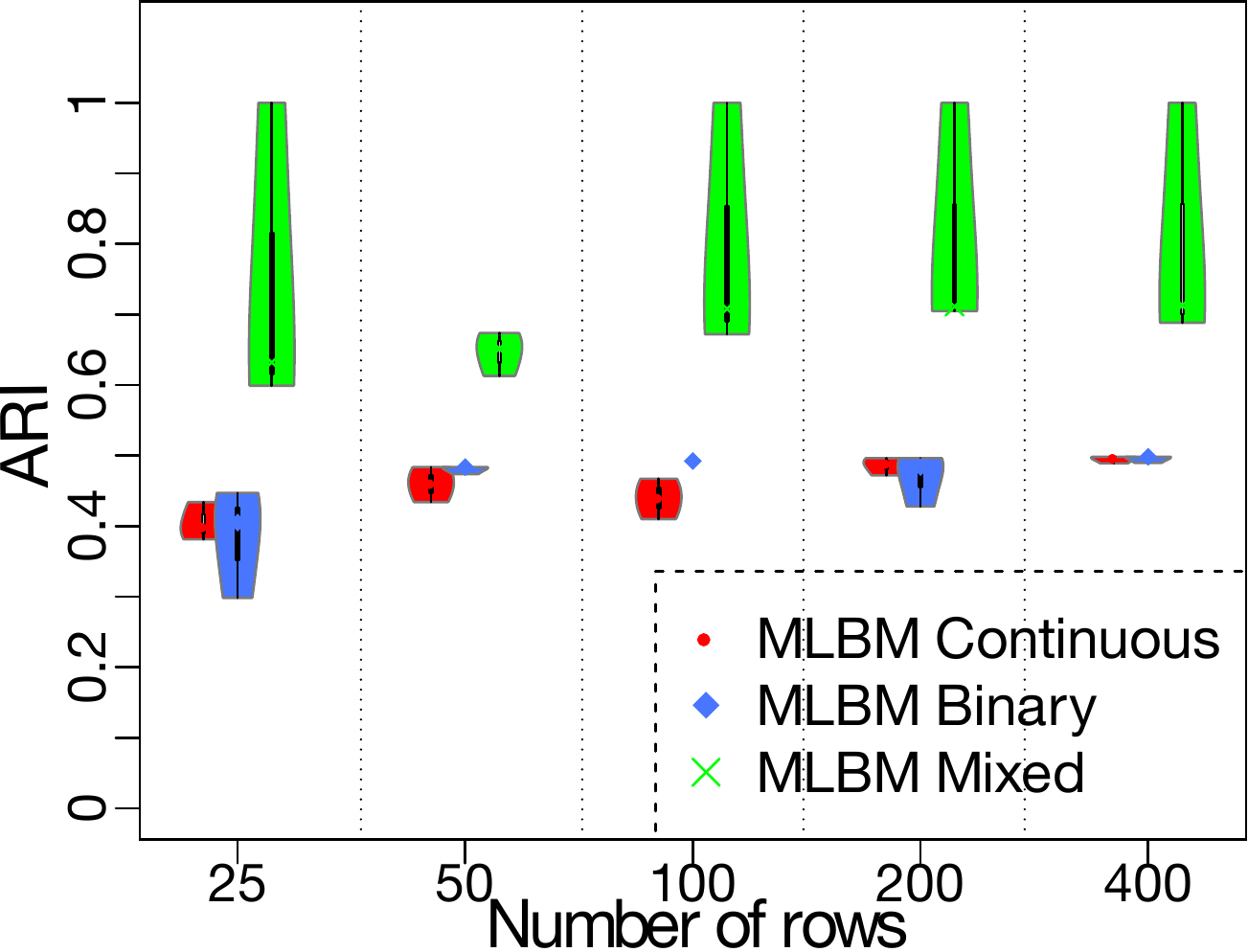}\label{p9_experiences1lowR}}
\quad
\subfloat[ARI of rows, Medium confusion]{\includegraphics[width=0.31\textwidth]{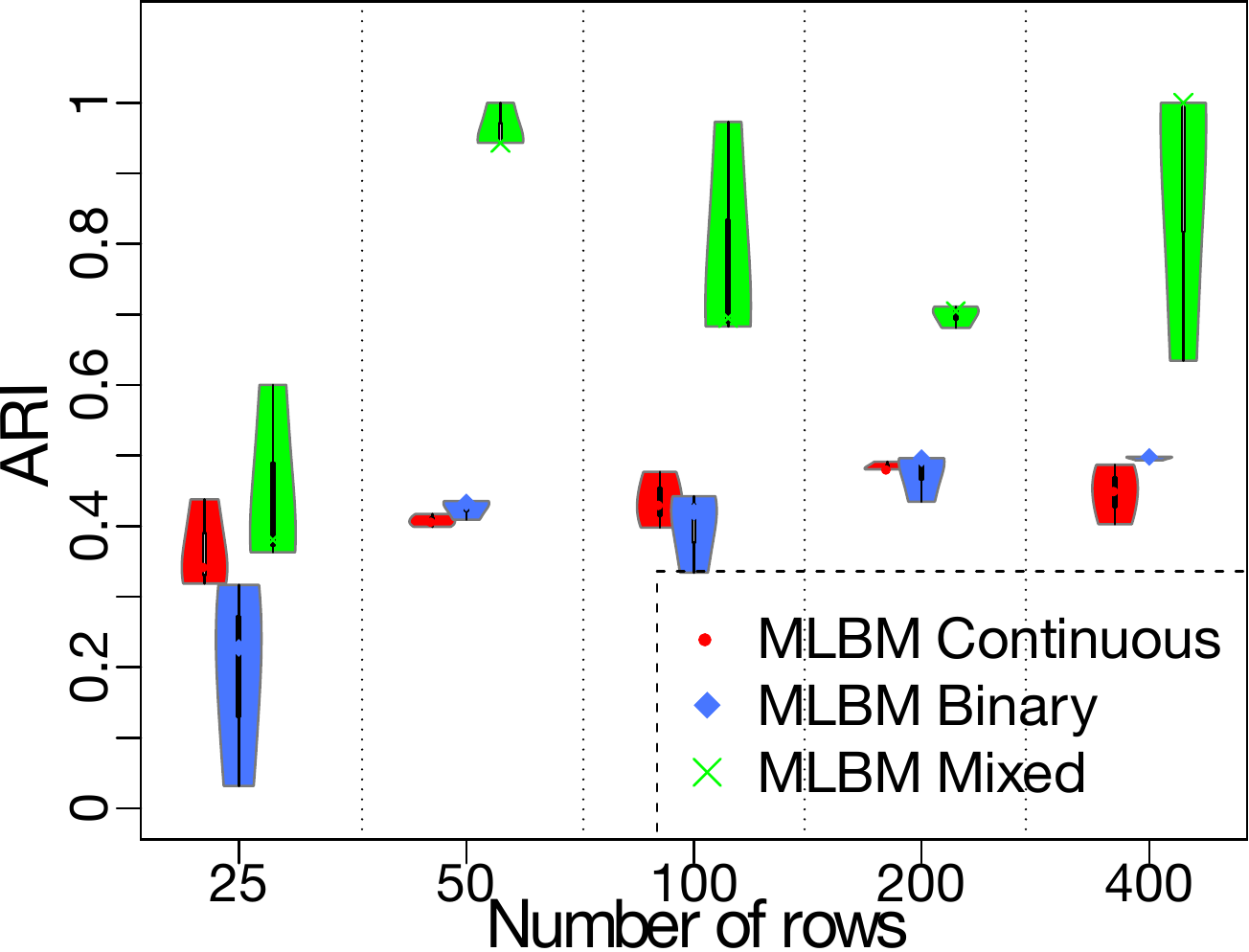}\label{p9_experiences1mediumR}}
\quad
\subfloat[ARI of rows, High confusion]{\includegraphics[width=0.31\textwidth]{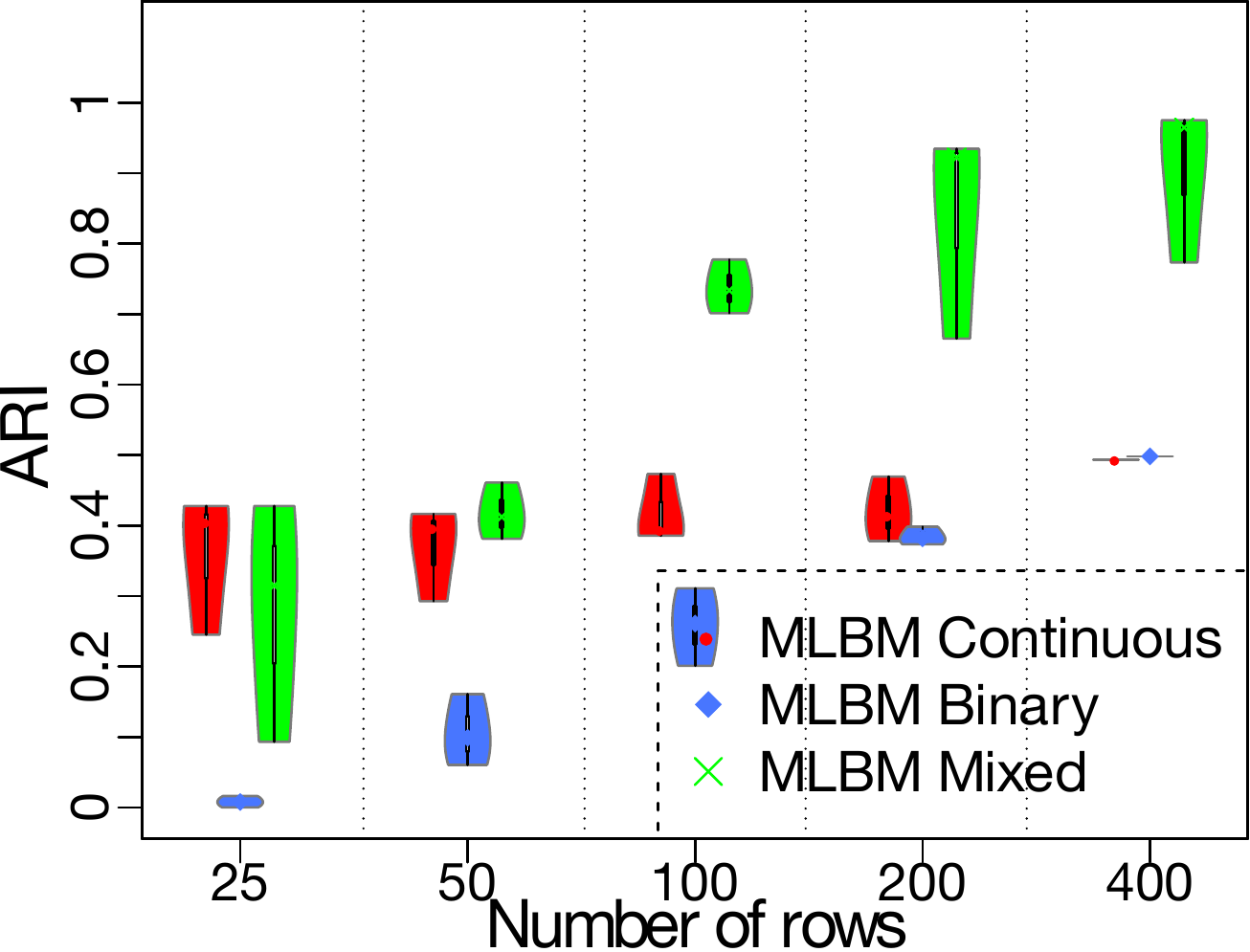}\label{p9_experiences1highR}}
\quad
\subfloat[ARI of columns, Low confusion]{\includegraphics[ width=0.31\textwidth]{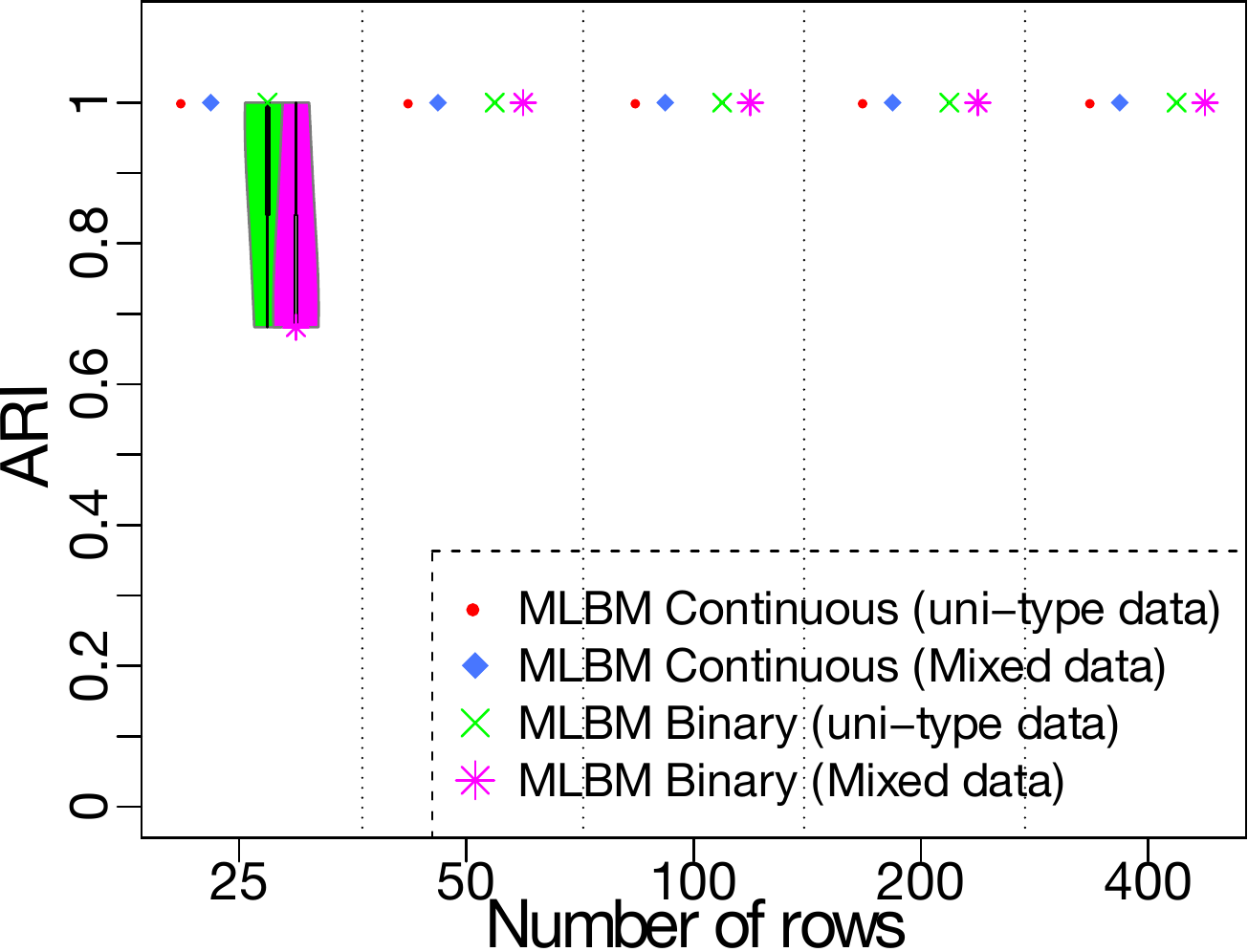}\label{p9_experiences1lowC}}
\quad
\subfloat[ARI of columns, Medium confusion]{\includegraphics[width=0.31\textwidth]{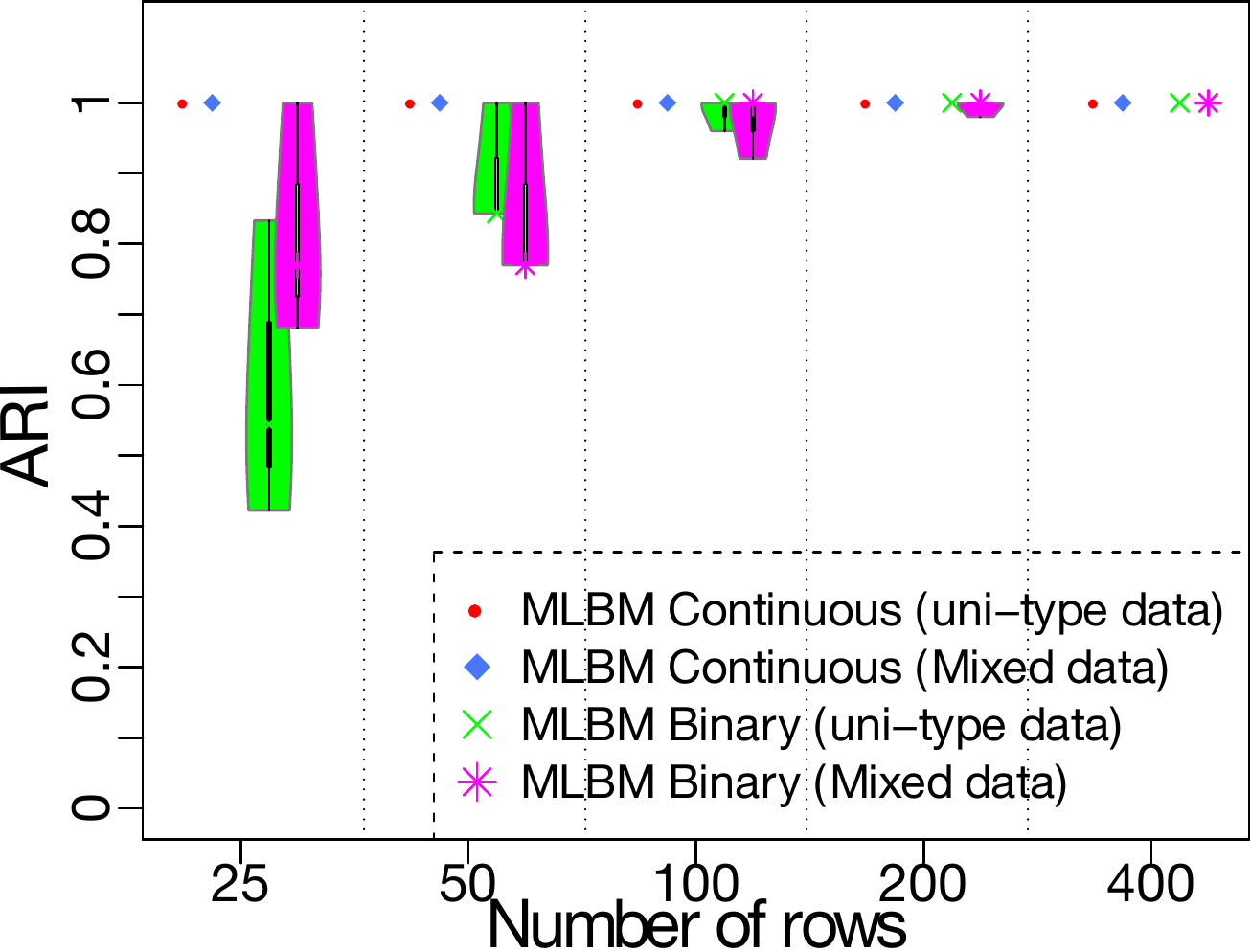}\label{p9_experiences1mediumC}}
\quad
\subfloat[ARI of columns, High confusion]{\includegraphics[width=0.31\textwidth]{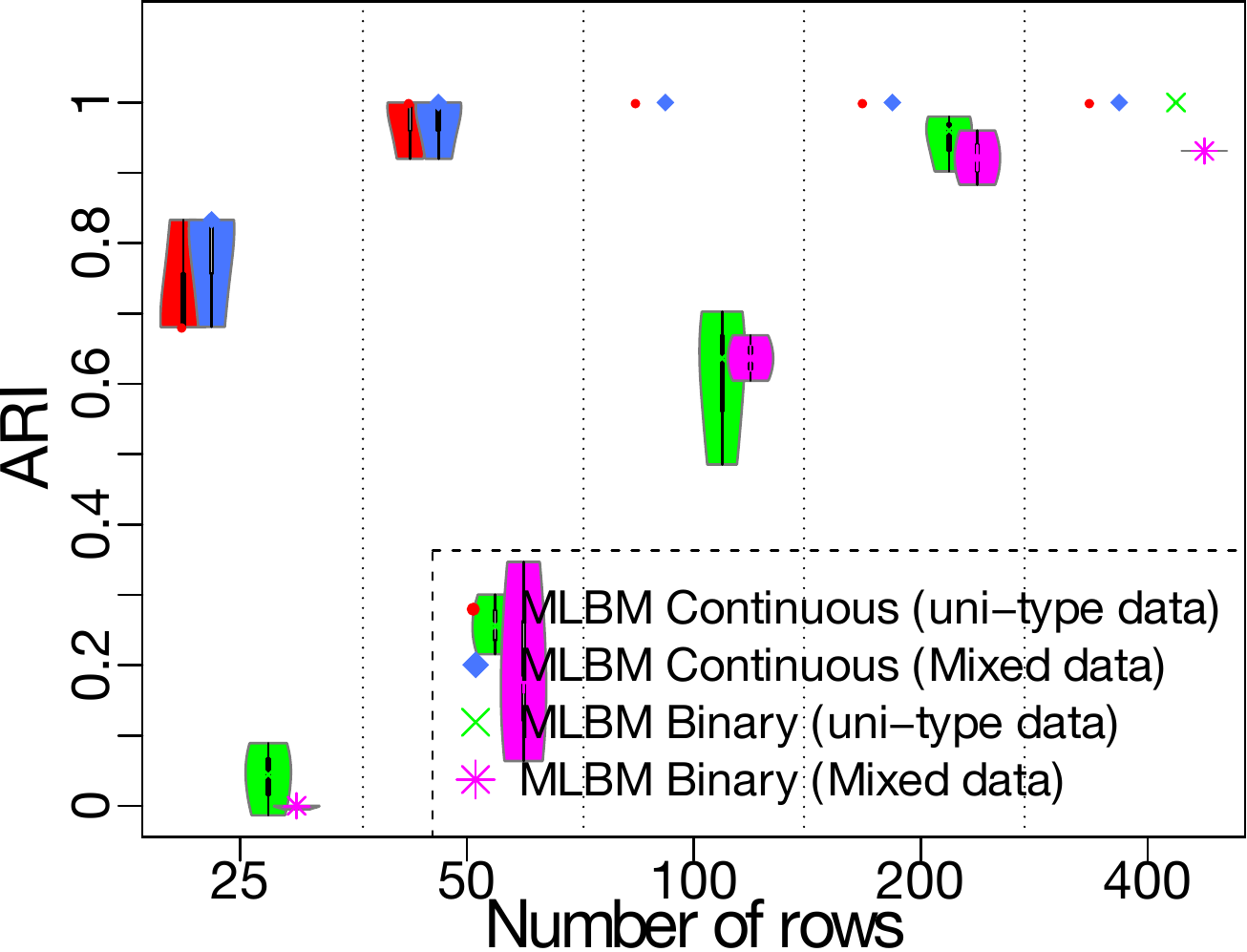}\label{p9_experiences1highC}}
\quad
\caption{{\it {\small{First experiment: ARI of rows and ARI of columns (in the y-axis) in the continuous, binary and mixed data.}}}}\label{p9_experiences1}
\end{figure}

Given such mixed data, the correspondence between the continuous and binary partitions is virtually null. This leaves the choice open for interpreting either the continuous co-clusters or the binary ones. Our approach proposes to use the full data and provides co-clusters for which the accuracy of the row clusters is at least as good as the best of the two choices. Furthermore,  mixed data co-clustering significantly  improves the accuracy of the retrieved partition in the majority of the studied cases (Figure~\ref{p9_experiences1}).

From Figure~\ref{p9_experiences1}, it is clear that, regardless of the level of overlap between the distributions and regardless of the size of the matrix, co-clustering the mixed data,  instead of separately co-clustering the continuous and the binary parts, improves significantly the quality of the obtained row partition (see Figures~\ref{p9_experiences1lowR}, \ref{p9_experiences1mediumR}, and \ref{p9_experiences1highR}). In fact, in the worst case scenarios, mixed data co-clustering  provides ARI of rows that are at least as good as the best ARI results when  performing uni-type data analysis. On the other hand, the adjusted Rand indexes of columns do not necessarily improve significantly (in some cases, it does), which is expected  because the configuration is set so that the clusters of columns are separable using uni-type data and the mixed analysis would not improve the performance of the clustering of columns (independence between the two data types in terms of column clusters). With respect to the data size and the level of overlap between the distributions, we notice the following. 

\begin{itemize}
\item Influence of the data size: as the data size increases, the quantity of the data units used by the optimization algorithm increases, which facilitates the convergence of the algorithms to the true underlying distributions. This effect can be observed from the ARI values, shown in Figure~\ref{p9_experiences1}, and mainly in the case of binary and mixed data.

\item Influence of the level of confusion: as expected, when the level of confusion between the distributions increases, it becomes harder to recover the exact true partition of rows. This effect is particularly visible in Figures~\ref{p9_experiences1highC} and \ref{p9_experiences1highR}, where the high level of confusion makes the separation of the clusters  difficult in the case of binary (and consequently mixed) data and particularly in small matrices. 
\end{itemize}

To summarize, the joint co-clustering of the continuous and binary variables of the simulated data sets enables us to use the full data and obtain considerably better accuracy, compared to an independent analysis by data type. The results of the co-clustering (both uni-type and mixed) are at their best when the level of confusion is low or the data matrix is big. With respect to the level of confusion, this behavior is expected since the true structure of the data is well separable. In fact, the level of confusion simulates the overlap between the distributions. Therefore, the higher the overlap, the data will contain more observations with relatively equal probabilities to belong to either of the distributions. Hence, a decrease in the accuracy of the clustering as it is measured over all the observations. The effect of bigger matrices can be explained by the fact that the more data is present, the more iterations are required by the algorithm which improves the quality of the estimated parameters.  

However, this is a well known phenomenon in the standard the standard LBM context. For example, in \cite{govaert2013}, the authors note that given the same number of co-clusters in the data, the classification error rate  depends not only on the parameters but also on the size of the data matrix (the Bayes classification risk decreases with the size of the data). Also, \cite{Matias2015} show that, when the estimated parameters converge to the true parameters, the recovered partitions will converge to the true partitions when the size of the data becomes sufficiently large. Using the mixed data latent block model, the accuracy of the estimated parameters is remarkable which reinforces the hypothesis that, as in the standard latent block models, given enough data, our approach would converge to the true partitions. Table~\ref{p9_Estimated_SG} shows examples of the estimated parameters using the mixed data latent block model MLBM on the data containing $100$ rows.

\begin{table}
\begin{center}
\scalebox{0.8}{
\begin{tabular}{|lc}
\hline
Low confusion &
\begin{tabular}{|c|c|c|}
\hline
(True $\mu$, estimated $\mu$) & (True $\sigma$, estimated $\sigma$) & (True $\alpha$, estimated $\alpha$)
\\
\hline
\begin{tabular}{ccc}
& $Jc_1$      & $Jc_2$\\
$I_1$ & (2,  2.001) & (1, 1.004)\\
$I_4$  &(2, 1.992) & (2, 1.013)\\
$I_2$ &(2, 2.000) & (2, 2.005)\\
$I_3$ &(2, 1.988) & (1, 0.975)\\
\end{tabular} & \begin{tabular}{cc}
$Jc_1$    & $Jc_2$\\
(0.25, 0.246) & (0.25, 0.255)\\
(0.25,  0.250) & (0.25, 0.239)\\
(0.25,  0.251) & (0.25, 0.248)\\
(0.25, 0.266) & (0.25, 0.259)\\
\end{tabular} & \begin{tabular}{cc}
$Jd_2$    & $Jd_1$\\
(0.2, 0.208) & (0.8, 0.790)\\
(0.8, 0.804) & (0.8, 0.822)\\
(0.2, 0.496) & (0.8, 0.811)\\
(0.8, 0.797) & (0.8, 0.758)\\
\end{tabular}
\end{tabular}\\
\hline
Medium confusion &
\begin{tabular}{|c|c|c|}
\hline
(True $\mu$, estimated $\mu$) & (True $\sigma$, estimated $\sigma$) & (True $\alpha$, estimated $\alpha$)
\\
\hline

\begin{tabular}{ccc}
& $Jc_2$      & $Jc_1$\\
$I_2$ & (2, 1.981) & (2, 1.985)\\
$I_1$ & (1, 1.023) & (2, 1.995)\\
$I_4$& (2, 1.990) & (2, 2.010)\\
$I_3$ & (1, 1.013) & (2, 1.962)\\
\end{tabular} & \begin{tabular}{cc}
 $Jc_2$      & $Jc_1$\\
(0.50, 0.520) & (0.50, 0.497)\\
(0.50, 0.502) & (0.50, 0.493)\\
(0.50, 0.505) & (0.50, 0.505)\\
(0.50, 0.501) & (0.50, 0.504)\\
\end{tabular} & \begin{tabular}{cc}
 $Jd_1$      & $Jd_2$\\
(0.7, 0.718) & (0.3, 0.297)\\
(0.7, 0.695) & (0.3, 0.315)\\
(0.7, 0.704) & (0.7, 0.685)\\
(0.7, 0.700) & (0.7, 0.674)\\
\end{tabular}
\end{tabular}\\
\hline
High confusion &
\begin{tabular}{|c|c|c|}
\hline
(True $\mu$, estimated $\mu$) & (True $\sigma$, estimated $\sigma$) & (True $\alpha$, estimated $\alpha$)
\\
\hline
\begin{tabular}{ccc}
& $Jc_2$      & $Jc_1$\\
$I_2$&  (2, 1.994) &(2, 2.042)\\
$I_1$ & (1, 0.998) &(2, 1.987)\\
$I_4$ & (2, 2.005) &(2, 2.008)\\
$I_3$ & (1, 1.012) &(2, 2.017)\\
\end{tabular} & \begin{tabular}{cc}
 $Jc_2$      & $Jc_1$\\
(1.00, 1.016) & (1.00, 0.989)\\
(1.00, 1.018) & (1.00, 0.979)\\
(1.00, 0.999) & (1.00, 1.001)\\
(1.00, 0.992) & (1.00, 1.008)\\
\end{tabular} & \begin{tabular}{cc}
 $Jd_2$      & $Jd_1$\\
(0.4, 0.351) & (0.6, 0.597)\\
(0.4, 0.399) & (0.6, 0.594)\\
(0.6, 0.591) & (0.6, 0.590)\\
(0.6, 0.631) & (0.6, 0.616)\\
\end{tabular}
\end{tabular}\\
\hline
\end{tabular}
}\caption{{\small{\it Examples of the estimated parameters for the $100$ rows data.}}}\label{p9_Estimated_SG}
\end{center}
\end{table}

\subsection{Second experiment}
\label{p9_expe2}
The objective of this experiment is to study the impact of the number of co-clusters, the size of the data and the level of confusion between the distributions.
\subsubsection{The data set}
 To study the influence of the number of co-clusters, the data sets are generated using the following parameters. 

\begin{itemize}
\item {\bf{The number of co-clusters}}: we choose three different partitions  $g\times(m_c+m_d)$ of the original data matrix: $2\times(2+2),\, 3\times(3+3),\,$ and $ 4\times(4+4)$.
\item {\bf{The size of the data}}: the size of the data is defined by the number of rows and the total number of columns. For this experiment, we choose the sizes $25$, $50$, $100$, $200$ and  $400$ for rows. For the number of columns, we distinguish two different configurations: 
\begin{itemize}
\item square matrices: the number of columns of each type is equal to the number of rows.
\item  rectangular matrices: we set the number of columns (of each type) to $5$, $10$, and  $20$.
\end{itemize} 

\item {\bf{The level of confusion}}: similarly to the first experiment, we consider three levels of overlap between the distributions: {\it Low} (with Gaussian means  $\mu \in \{p_1=1,\, p_2=2\}$, Gaussian standard deviations  $\sigma= 0.25$ and Bernoulli parameters   $\alpha \in\{p_1=0.2,\, p_2=0.8\}$), {\it  Medium} ($\mu \in \{p_1=1,\, p_2=2\}$, $\sigma= 0.5$ and $ \alpha \in\{p_1=0.3,\, p_2 = 0.7\}$) and  {\it High}  ($\mu \in \{p_1=1,\, p_2=2\}$, $\sigma= 1$ and  $ \alpha \in\{p_1=0.4, \,p_2=0.6\}$).  

\end{itemize}

\begin{table}[htbp]
\begin{center}
\scalebox{0.8}{
\begin{tabular}{c}
\begin{tabular}{lll}
\begin{tabular}{|l|l|l|l|}
  \hline
 $\mu$ or $\alpha$ & $J_1$ &$J_2$\\
  \hline
 $I_1$ & $p_1$ & $p_1$ \\
  \hline
  $I_2$ & $p_1$ & $p_2$ \\
  \hline
\end{tabular}& \begin{tabular}{|l|l|l|l|}
  \hline
 $\mu$ or $\alpha$  & $J_1$ &$J_2$ &$J_3$\\
  \hline
 $I_1$ & $p_1$ & $p_2$ & $p_1$\\
  \hline
  $I_2$ &$p_1$ & $p_2$ & $p_2$\\
      \hline
$I_3$ & $p_1$& $p_1$ & $p_1$ \\
  \hline
\end{tabular} &\begin{tabular}{|l|l|l|l|l|}
  \hline
$\mu$ or $ \alpha$  & $J_1$ &$J_2$ &$J_3$&$J_4$\\
  \hline
 $I_1$ & $p_2$&  $p_1$ &  $p_2$& $p_1$\\
  \hline
  $I_2$ & $p_2$ &  $p_1$ &  $p_2$& $p_2$\\
      \hline
$I_3$ &  $p_2$ &  $p_2$ &  $p_2$ & $p_2$\\
      \hline
$I_4$ &  $p_2$ &  $p_1$ &  $p_1$ & $p_1$\\
  \hline
\end{tabular}
\end{tabular}\\
\end{tabular}
}\caption{{\small{\it The true parameter specification with  $2\times(2+2)$, $3\times(3+3)$ and $4\times(4+4)$ co-clusters.}}}\label{p9_configurations2}
 \end{center}
\end{table}

The specification of the co-clusters and their configuration are shown in Table~\ref{p9_configurations2}. 
 Similarly to the first experiment, we generate 3 samples of each data configuration according to its parameters and we present the resulting ARI  in the form of violin plots. To present the co-clustering results, we distinguish between square matrices and rectangular ones. 

\subsubsection{The co-clustering results: square matrices}
Although each of the continuous and binary parts of the data can be sufficient to extract the underlying structure of the data,  we notice that, as in the first experiment, jointly co-clustering the continuous and binary data clearly  improves the performance of the co-clustering.

\begin{figure}[htbp]
\centering
\subfloat[ARI of rows, $2\times(2+2)$]{\includegraphics[ width=0.31\textwidth]{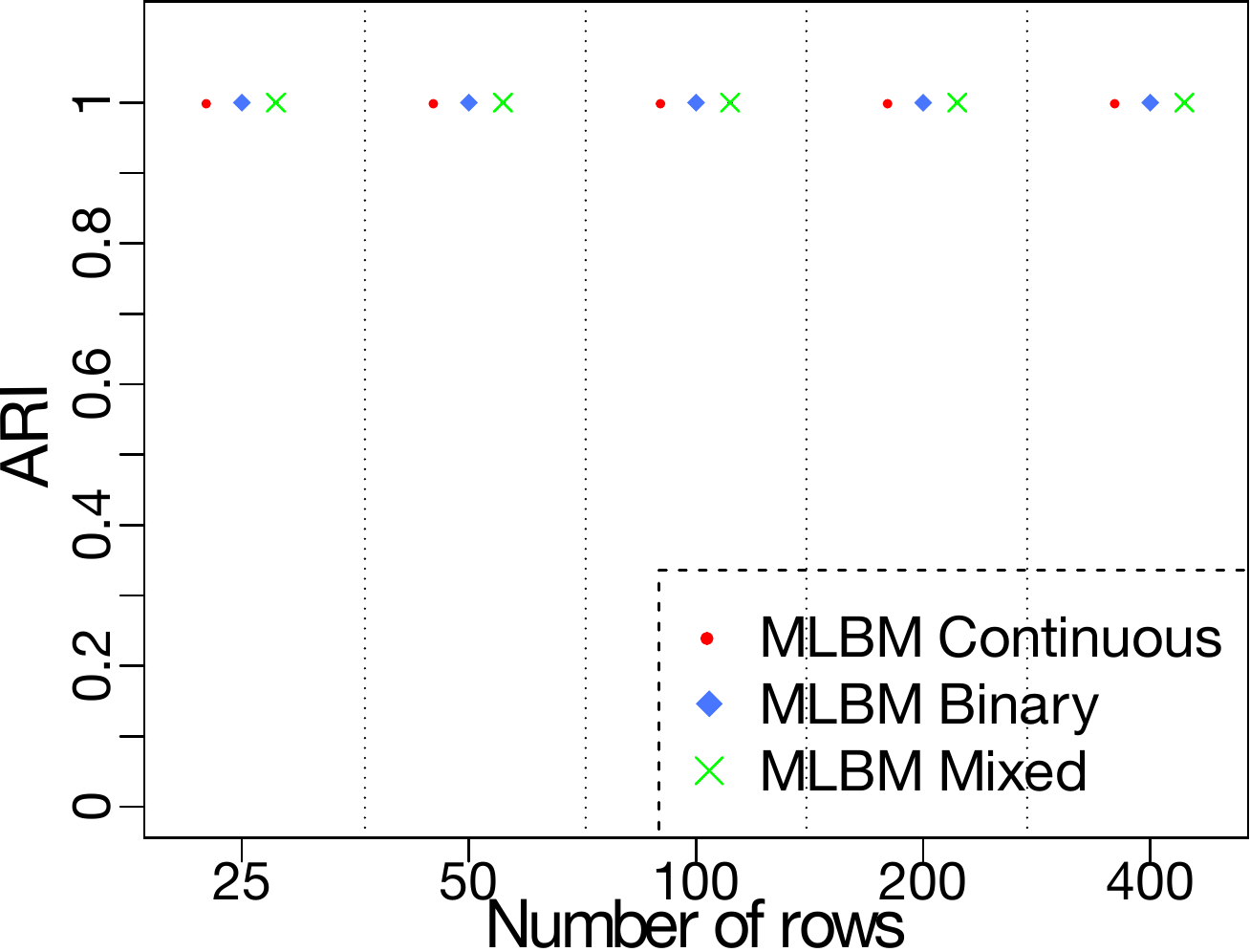}\label{p9_experiences21_low_222R}}
\quad 
\subfloat[ARI of rows, $3\times(3+3)$]{\includegraphics[width=0.31\textwidth]{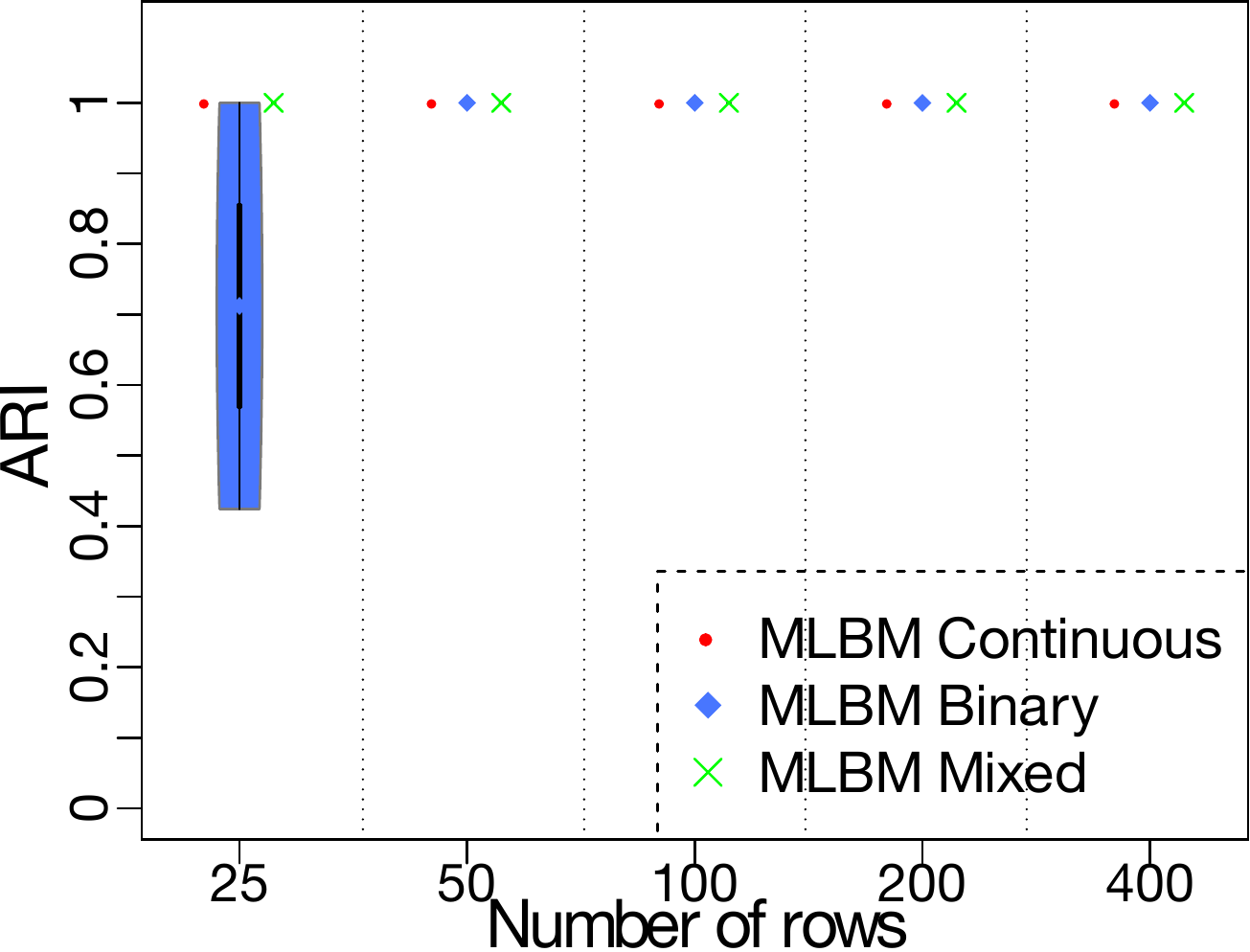}\label{p9_experiences21_low_333R}}
\quad
\subfloat[ARI of rows, $4\times(4+4)$]{\includegraphics[width=0.31\textwidth]{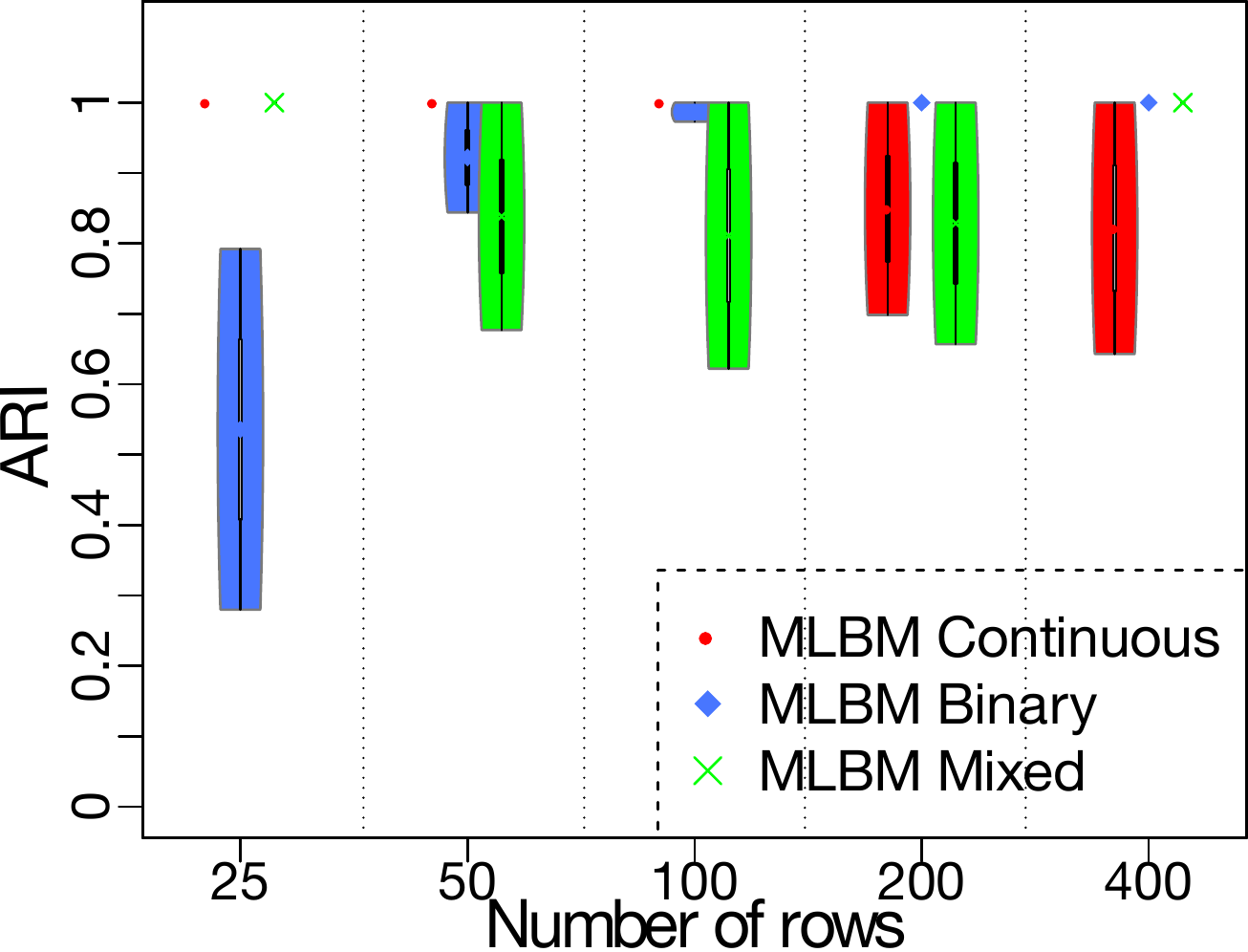}\label{p9_experiences21_low_444R}}
\quad
\subfloat[ARI of columns, $2\times(2+2)$]{\includegraphics[ width=0.31\textwidth]{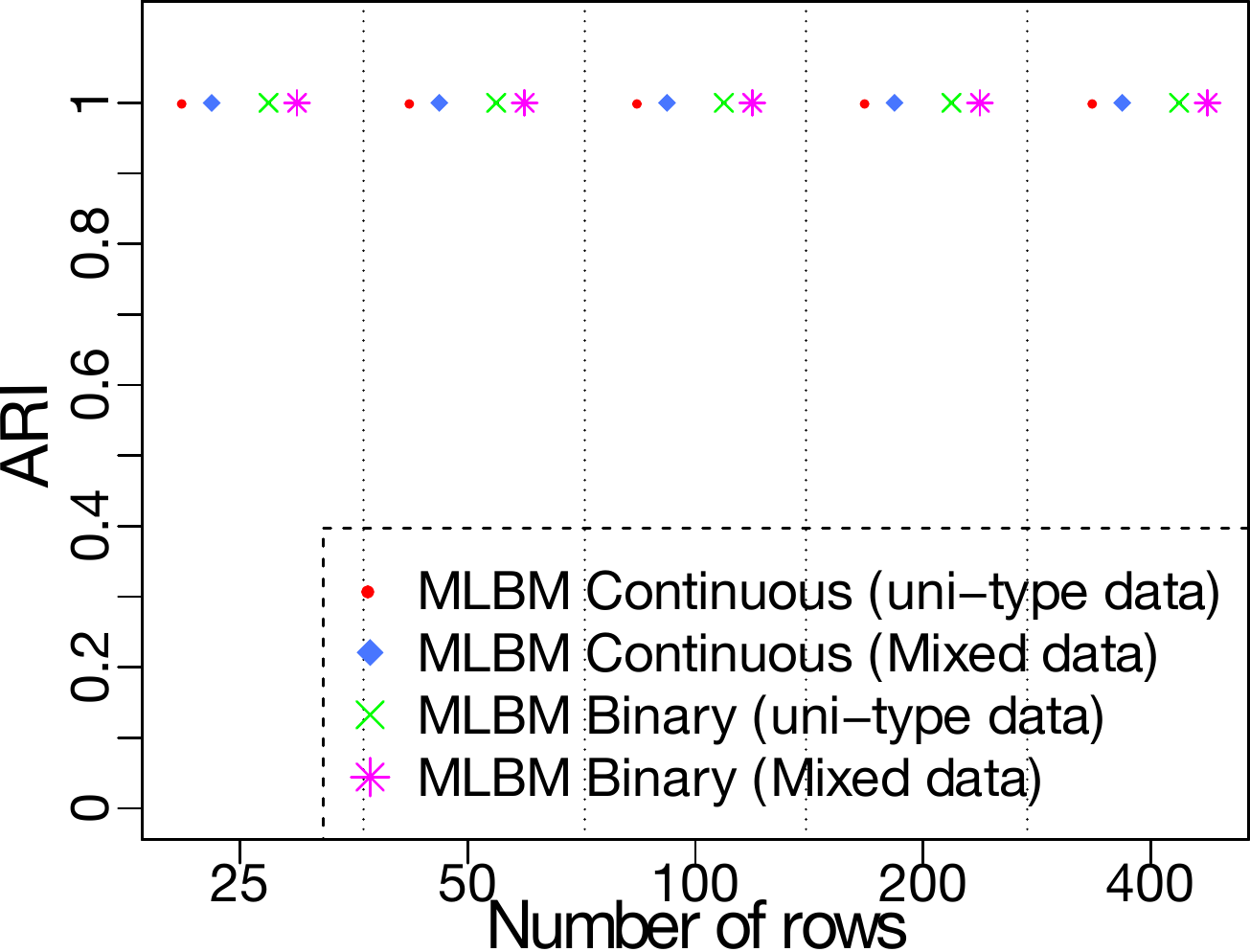}\label{p9_experiences21_low_222C}}
\quad
\subfloat[ARI of columns, $3\times(3+3)$]{\includegraphics[width=0.31\textwidth]{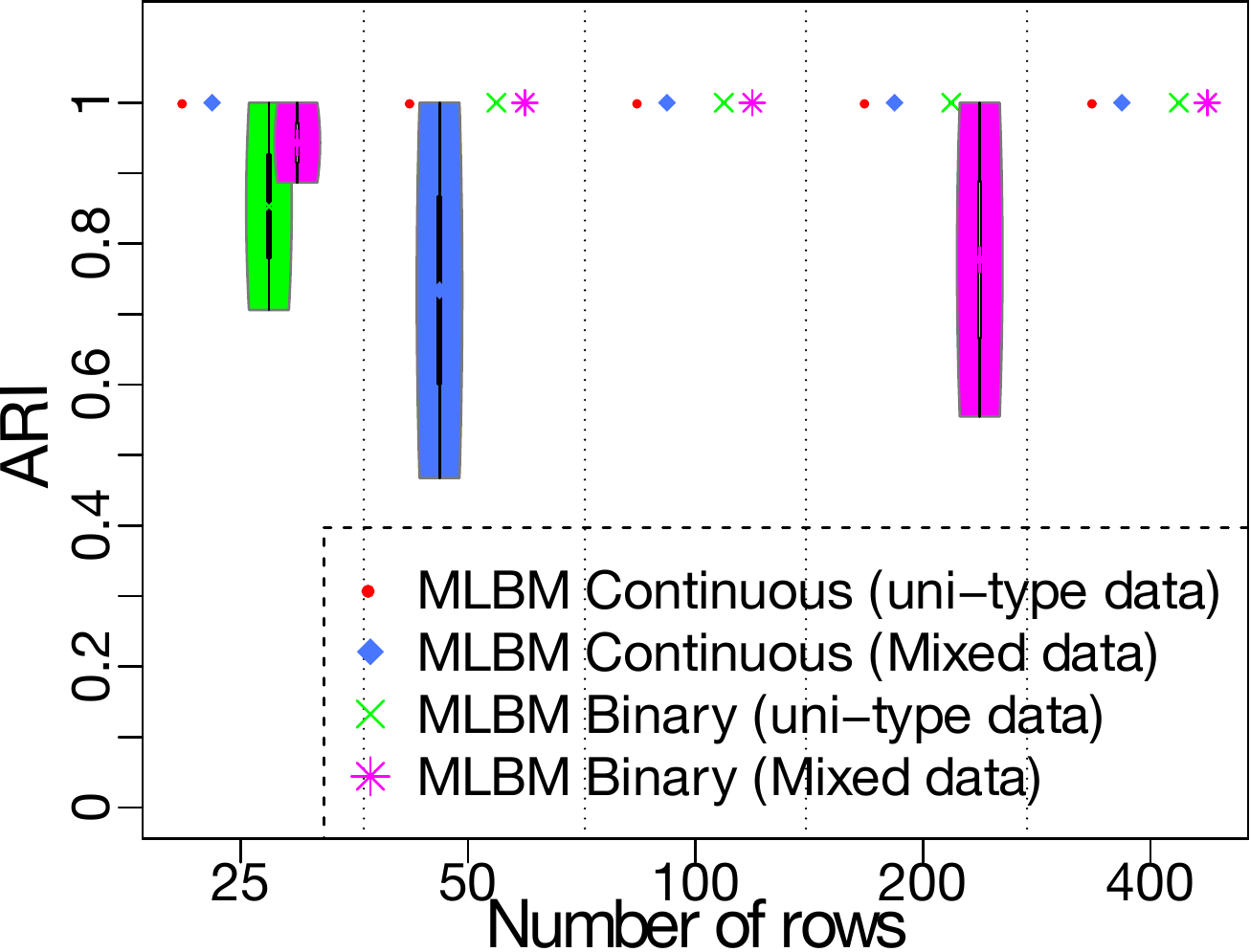}\label{p9_experiences21_low_333C}}
\quad
\subfloat[ARI of columns, $4\times(4+4)$]{\includegraphics[width=0.31\textwidth]{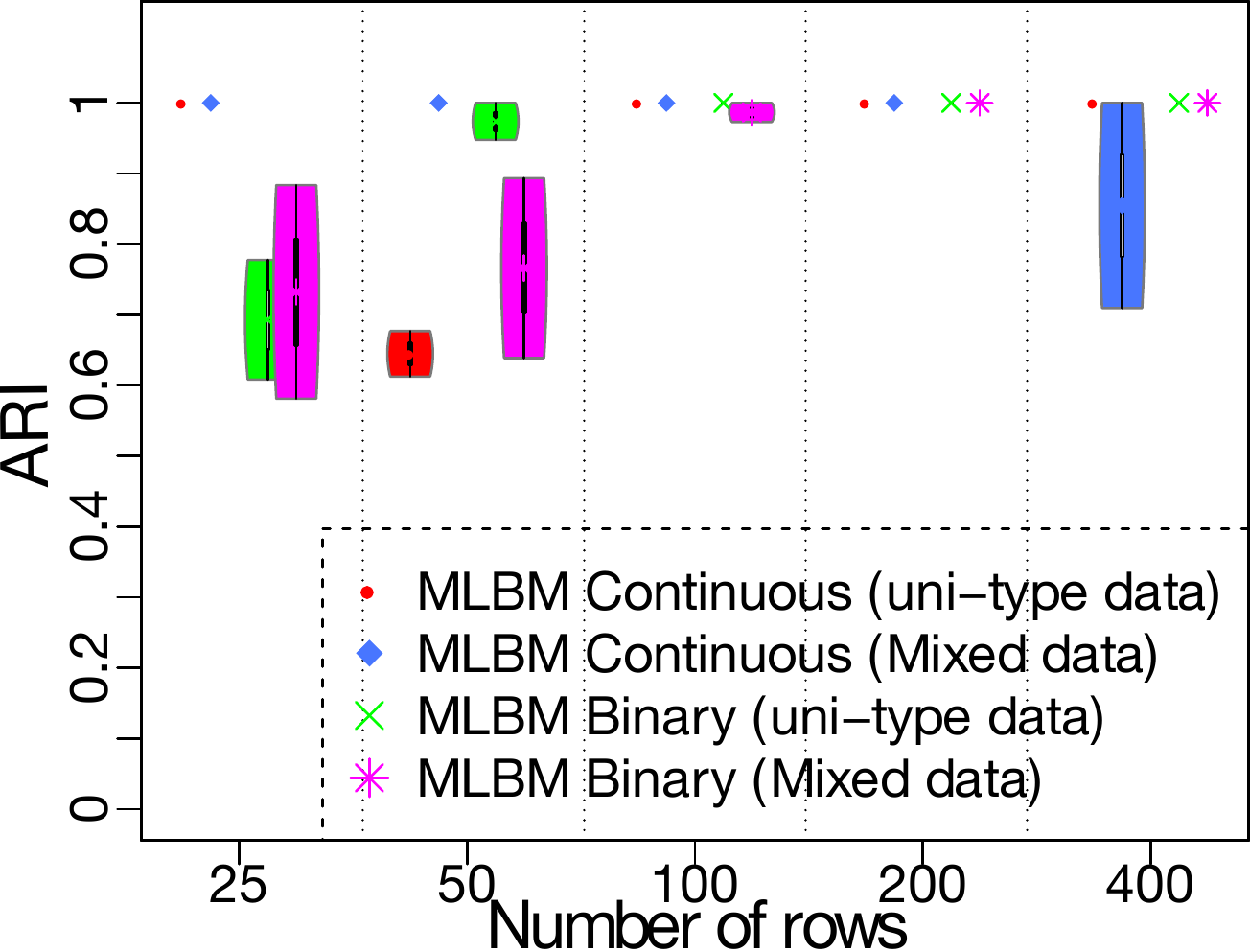}\label{p9_experiences21_low_444C}}
\quad
\caption{{\it {\small Second experiment (Low confusion): ARI of rows  and ARI of columns (in the y-axis) in the case of continuous, binary and mixed data.}}}\label{p9_experiences21}
\end{figure}

\begin{figure}[htbp]
\centering
\subfloat[ARI of rows, $2\times(2+2)$]{\includegraphics[width=0.31\textwidth]{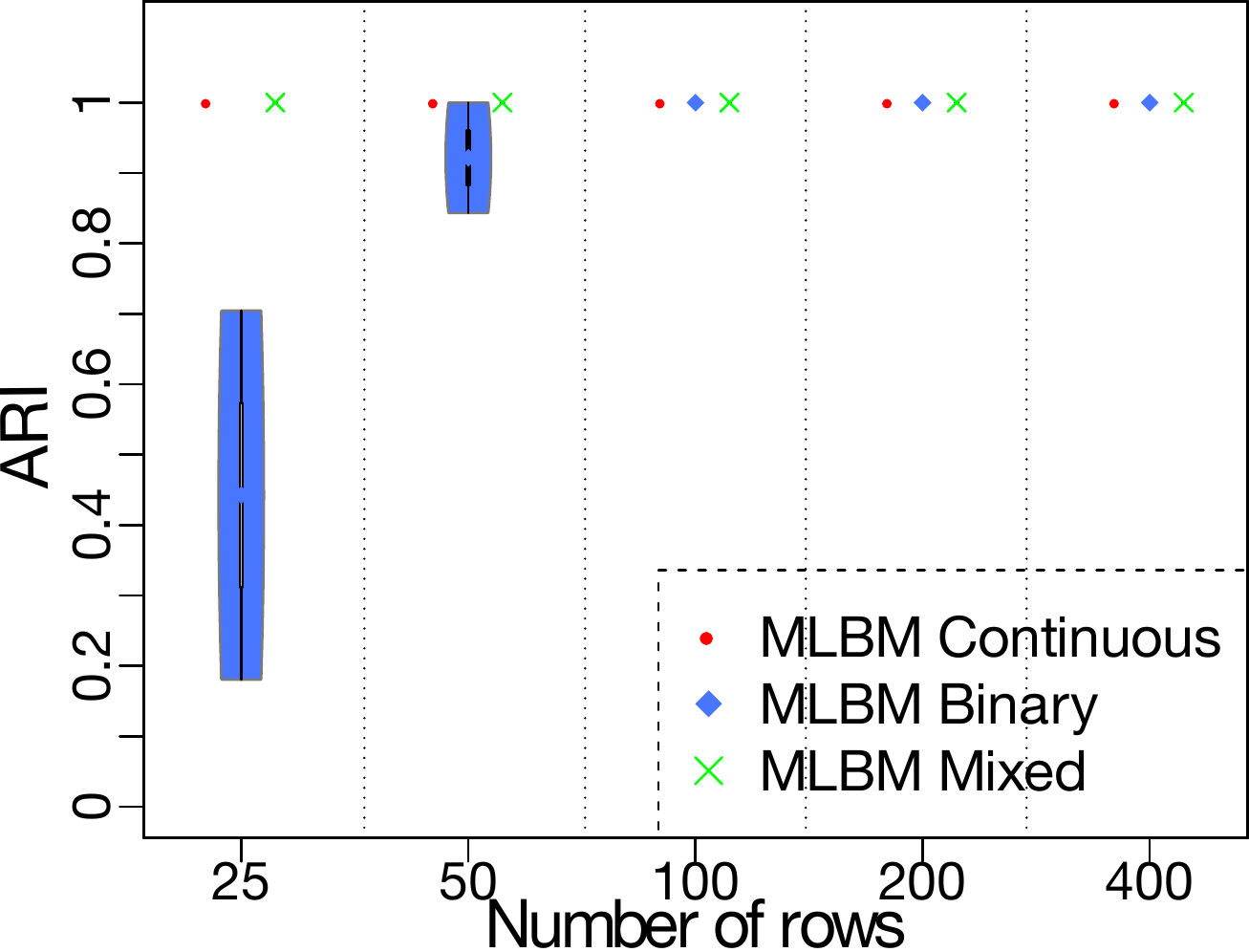}\label{p9_experiences22_medium_222R}}
\quad
\subfloat[ARI of rows, $3\times(3+3)$]{\includegraphics[width=0.31\textwidth]{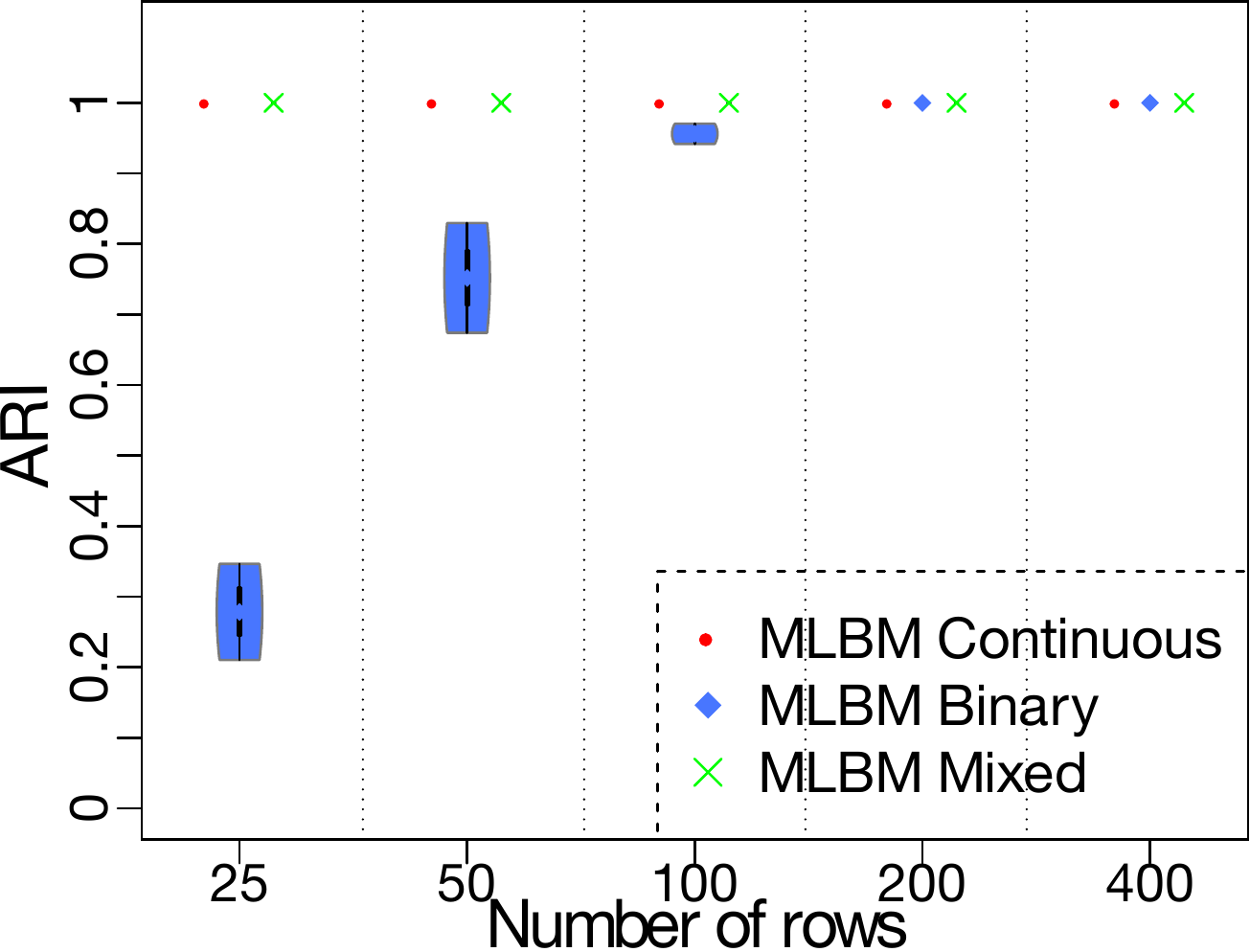}\label{p9_experiences22_medium_333R}}
\quad
\subfloat[ARI of rows, $4\times(4+4)$]{\includegraphics[width=0.31\textwidth]{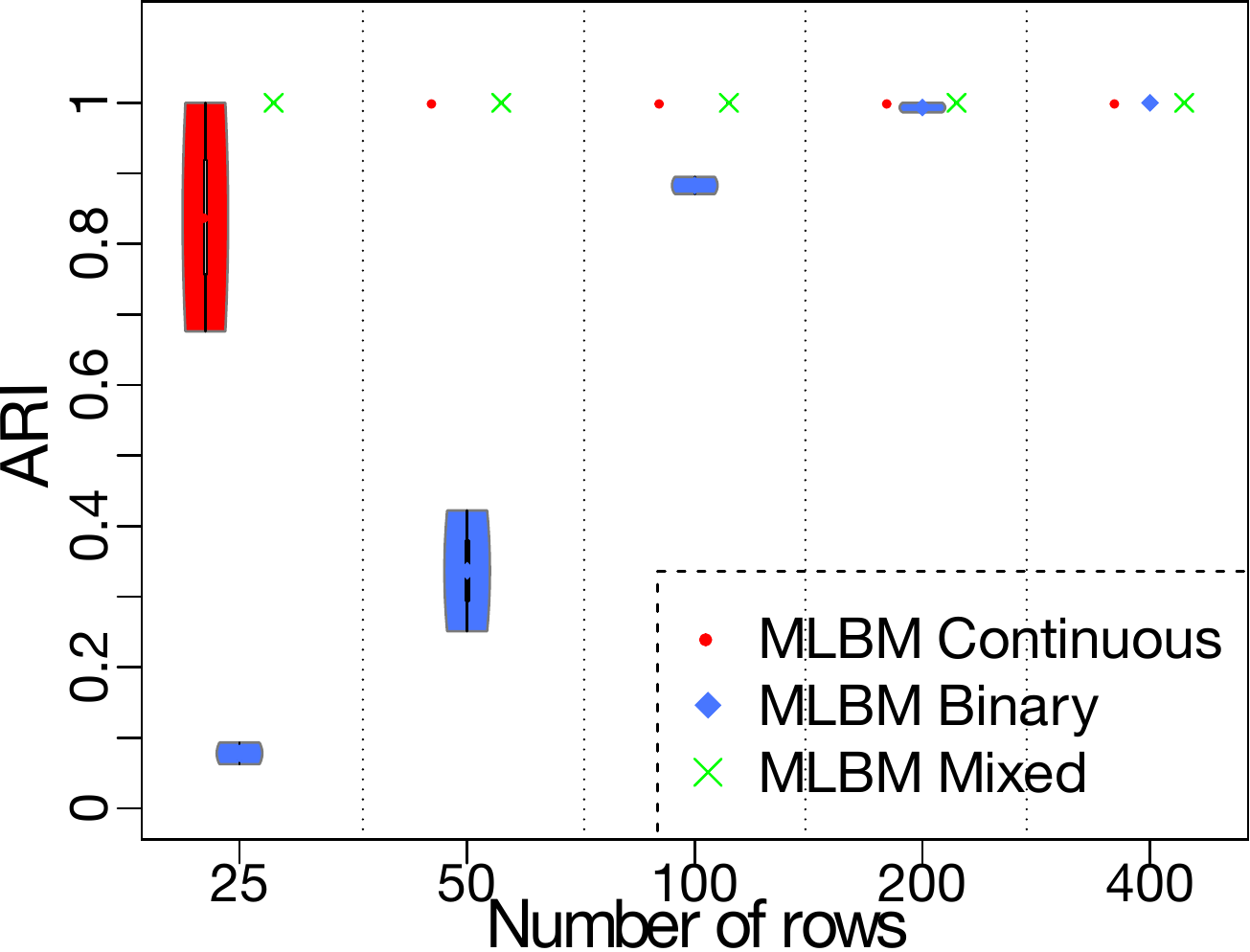}\label{p9_experiences22_medium_444R}}
\quad
\subfloat[ARI of columns, $2\times(2+2)$]{\includegraphics[width=0.31\textwidth]{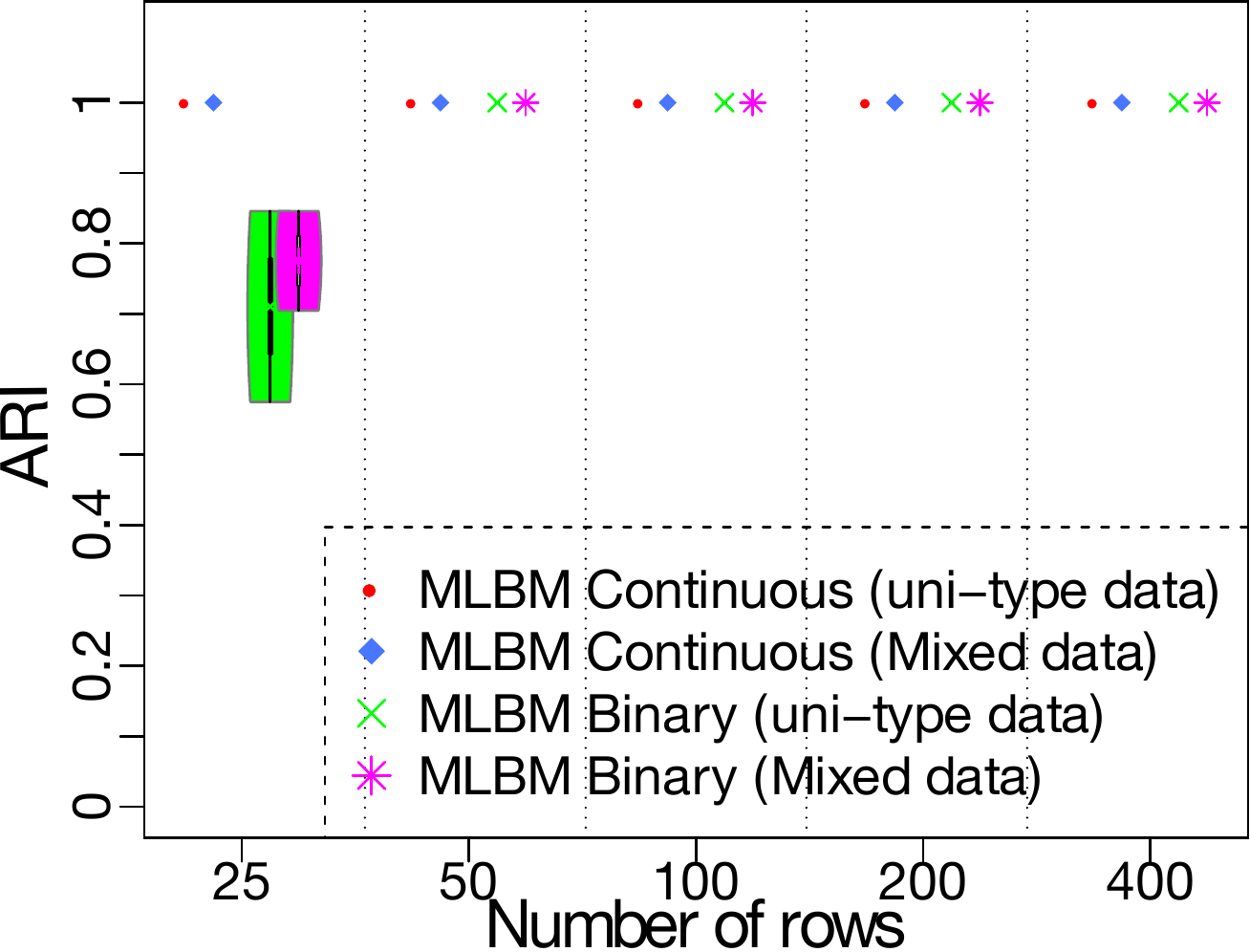}\label{p9_experiences22_medium_222C}}
\quad
\subfloat[ARI of columns, $3\times(3+3)$]{\includegraphics[width=0.31\textwidth]{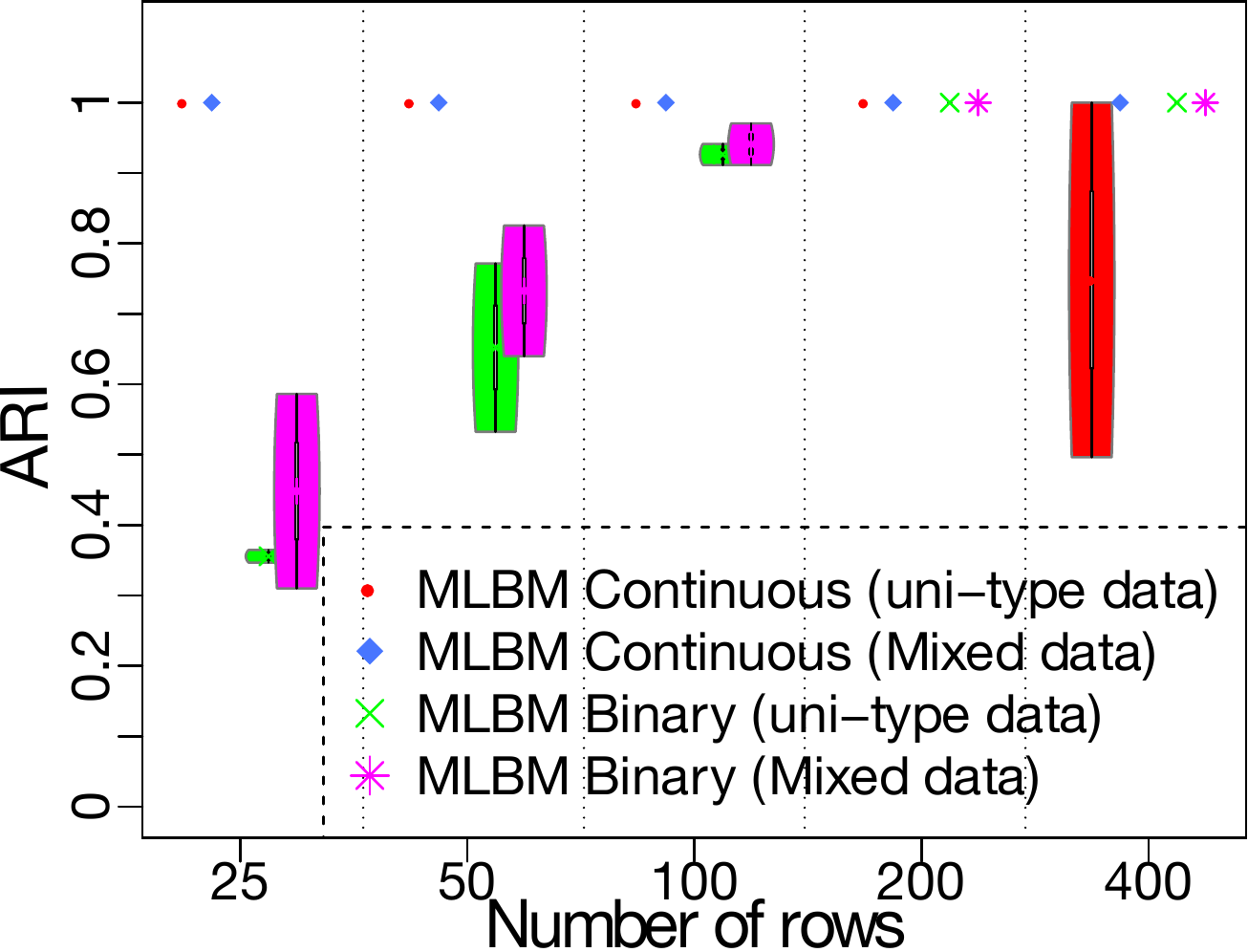}\label{p9_experiences22_medium_333C}}
\quad
\subfloat[ARI of columns, $4\times(4+4)$]{\includegraphics[width=0.31\textwidth]{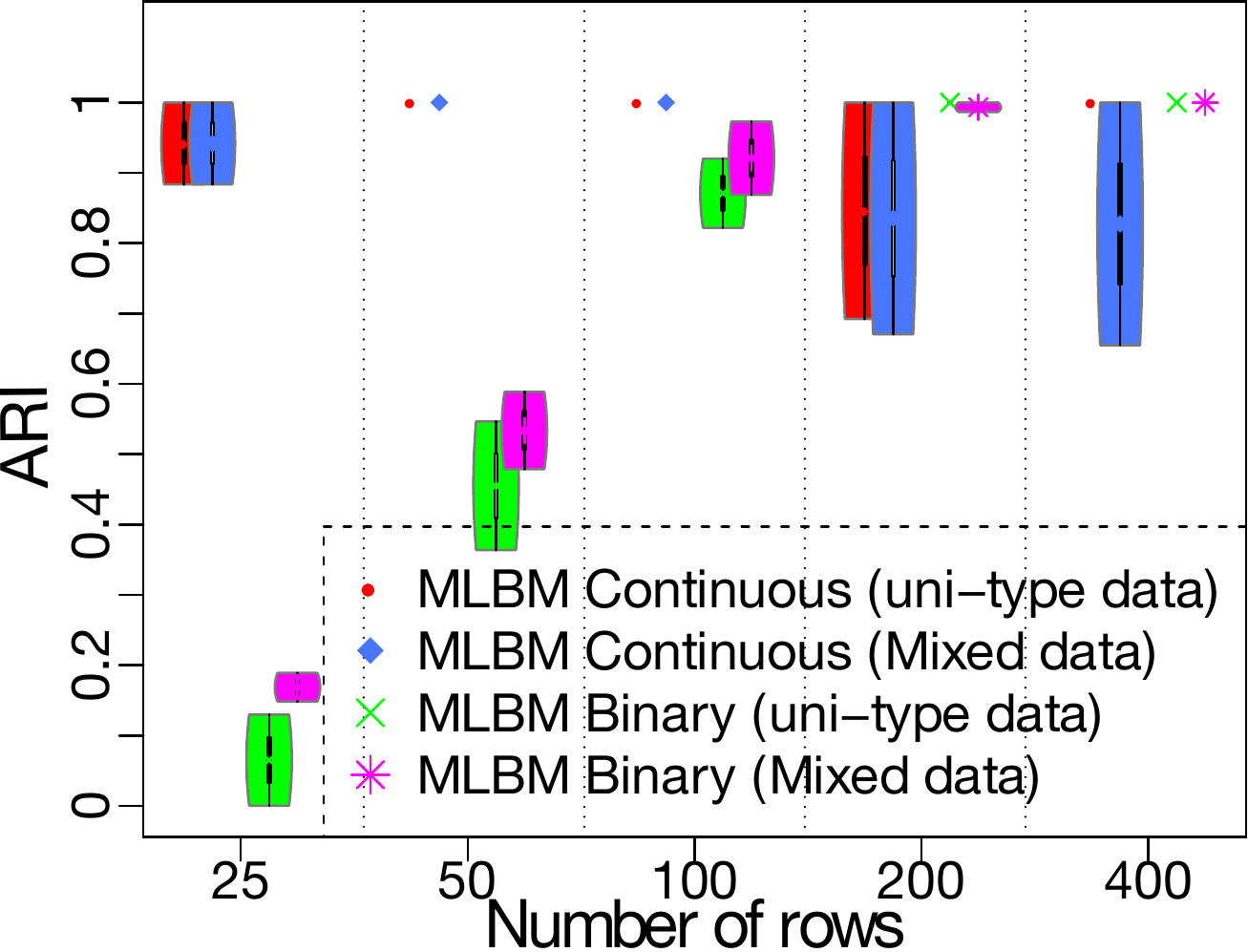}\label{p9_experiences22_medium_444C}}
\quad
\caption{{\it {\small Second experiment (Medium confusion): ARI of rows and ARI of columns (the y-axis) in the case of continuous, binary and mixed data.}}}\label{p9_experiences22}
\end{figure}

\begin{figure}[htbp]
\centering
\subfloat[ARI of rows, $2\times(2+2)$]{\includegraphics[width=0.31\textwidth]{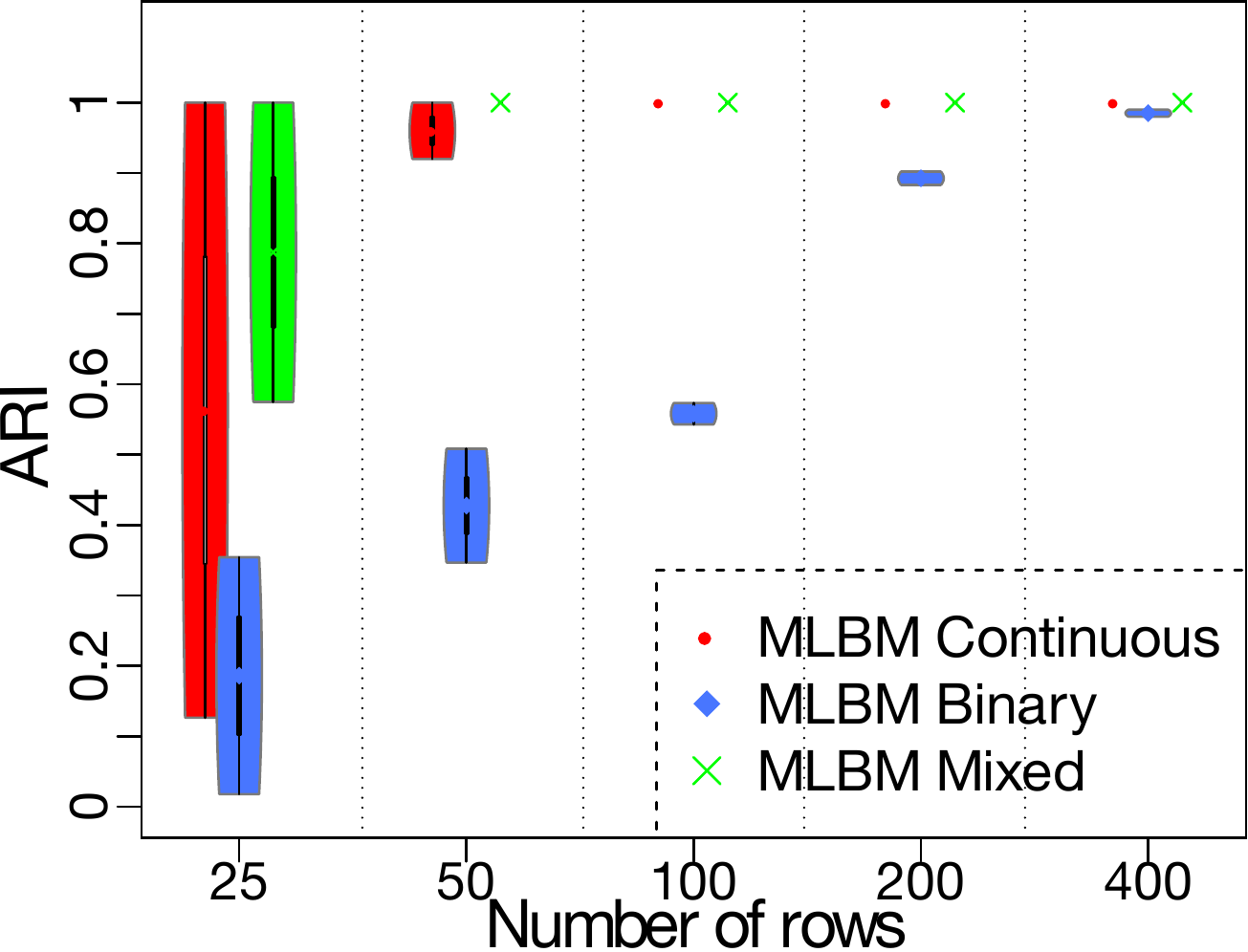}\label{p9_experiences23_high_222R}}
\quad
\subfloat[ARI of rows, $3\times(3+3)$]{\includegraphics[width=0.31\textwidth]{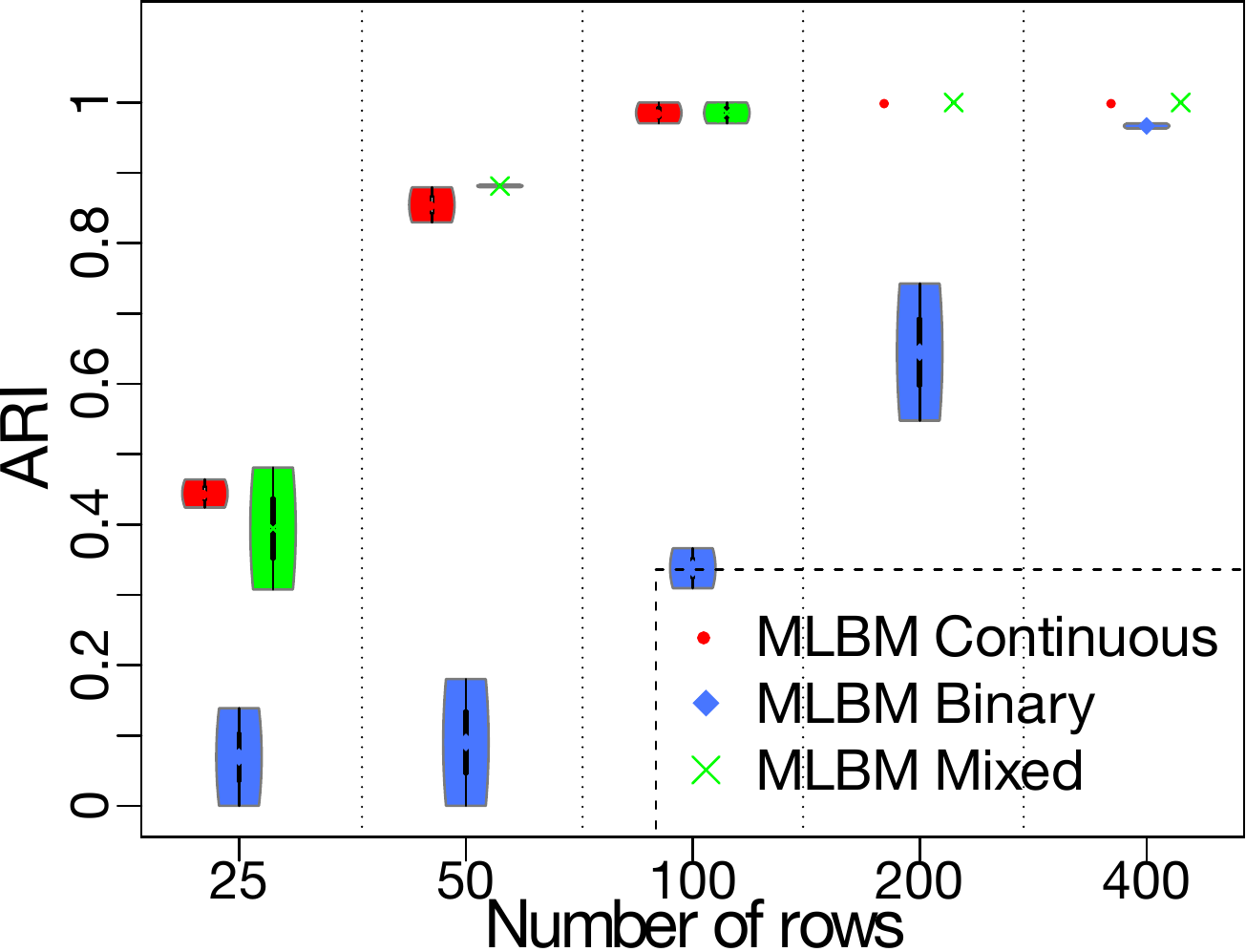}\label{p9_experiences23_high_333R}}
\quad 
\subfloat[ARI of rows, $4\times(4+4)$]{\includegraphics[width=0.31\textwidth]{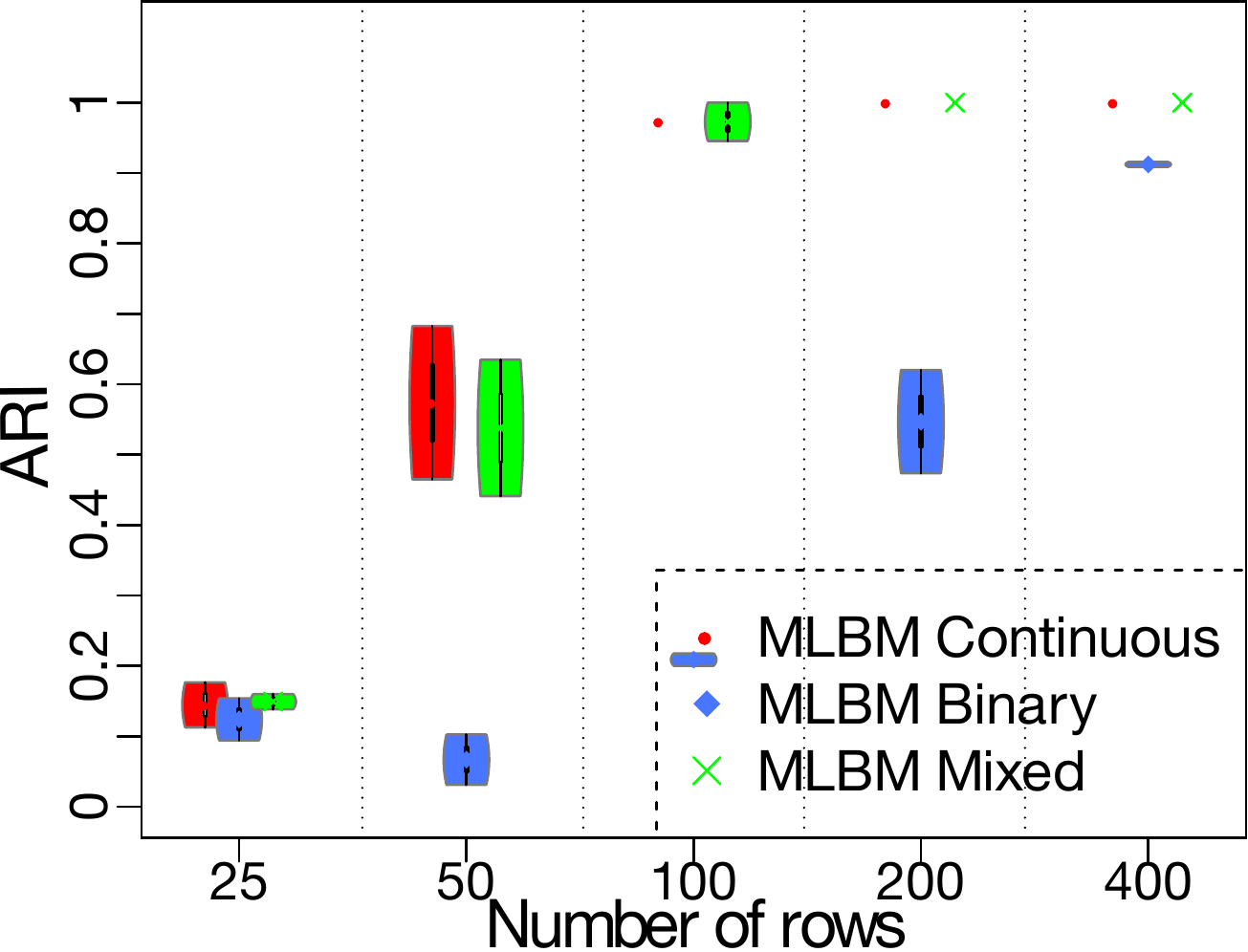}\label{p9_experiences23_high_444R}}
\quad
\subfloat[ARI of columns, $2\times(2+2)$]{\includegraphics[width=0.31\textwidth]{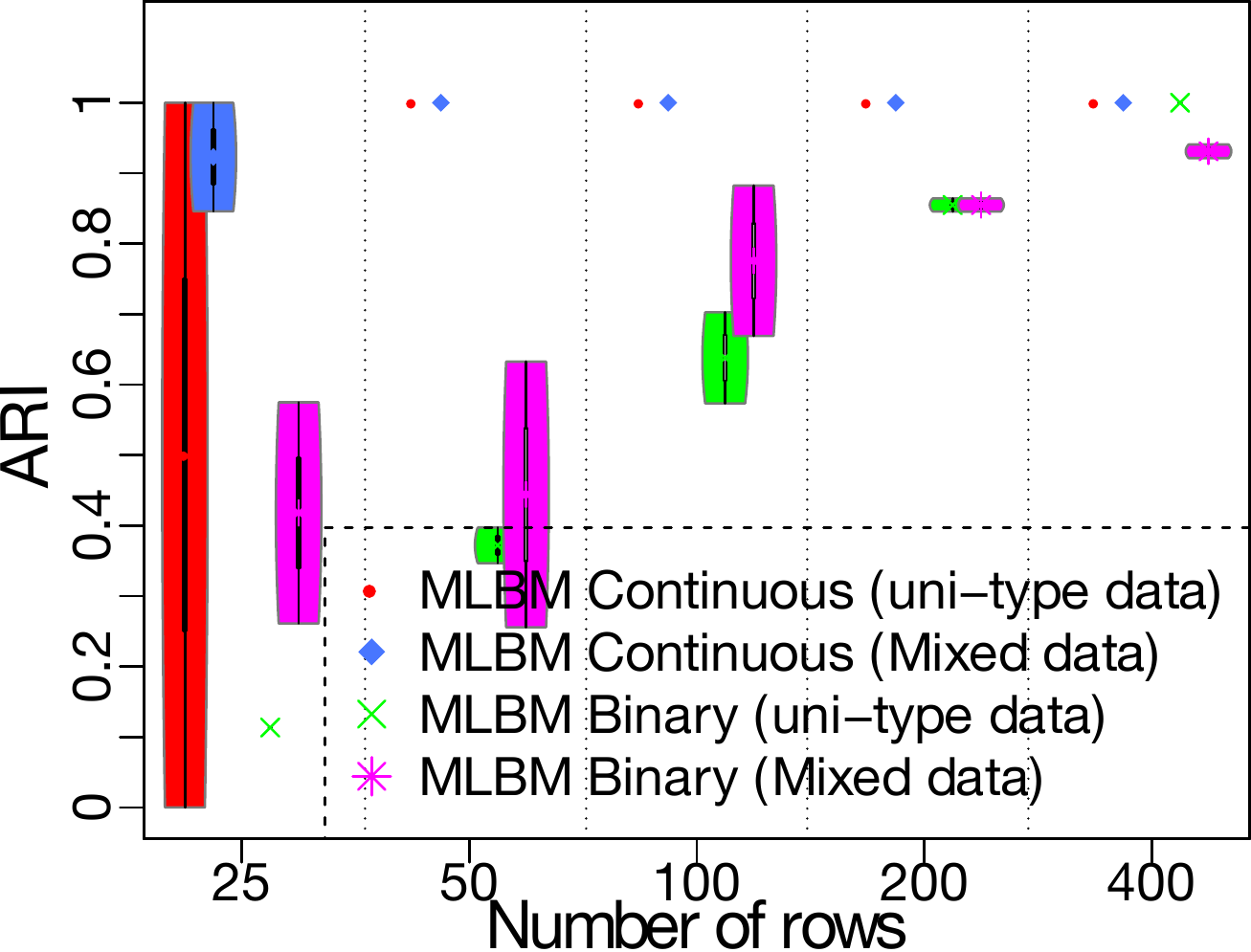}\label{p9_experiences23_high_222C}}
\quad 
\subfloat[ARI of columns, $3\times(3+3)$]{\includegraphics[width=0.31\textwidth]{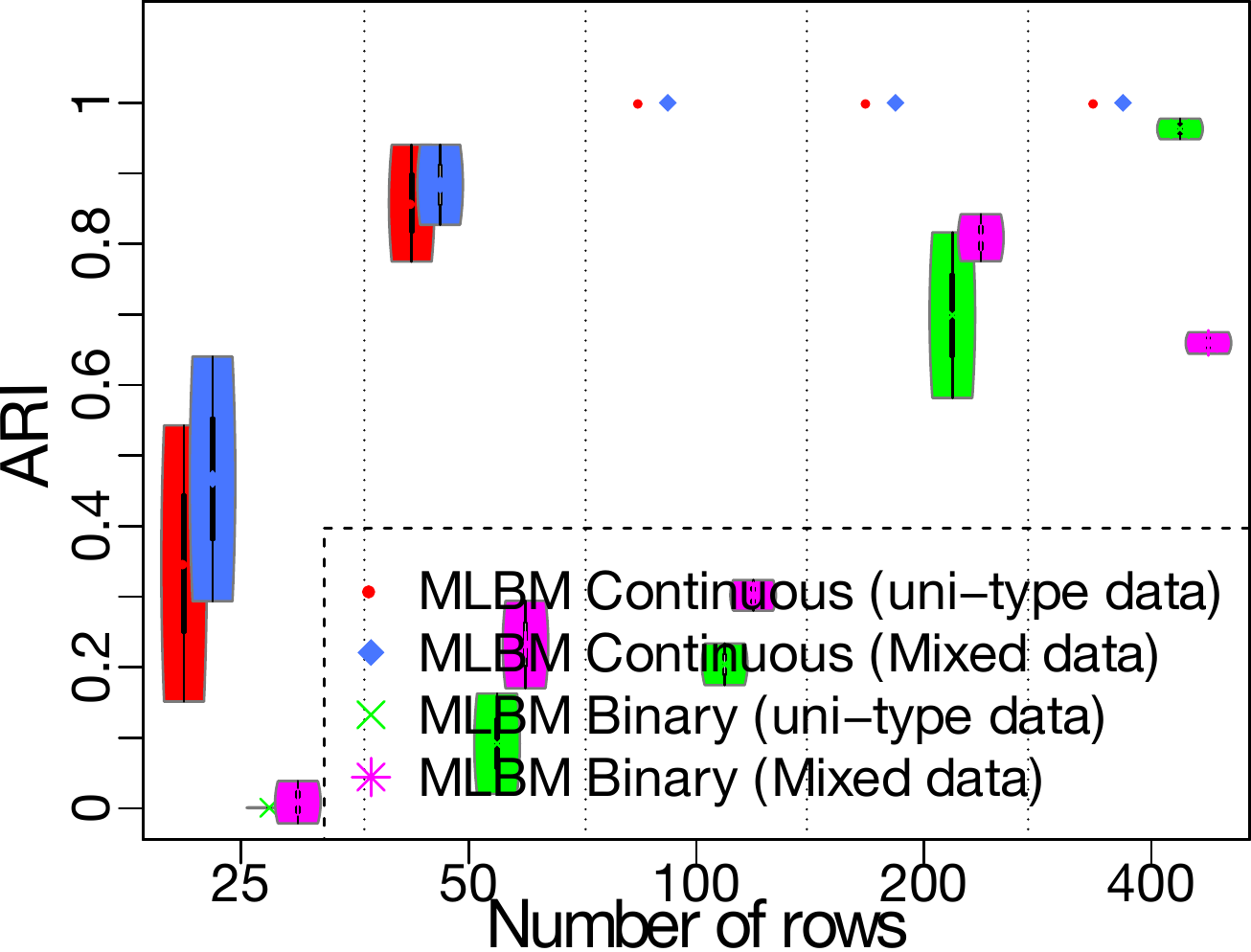}\label{p9_experiences23_high_333C}}
\quad
\subfloat[ARI of columns, $4\times(4+4)$]{\includegraphics[width=0.31\textwidth]{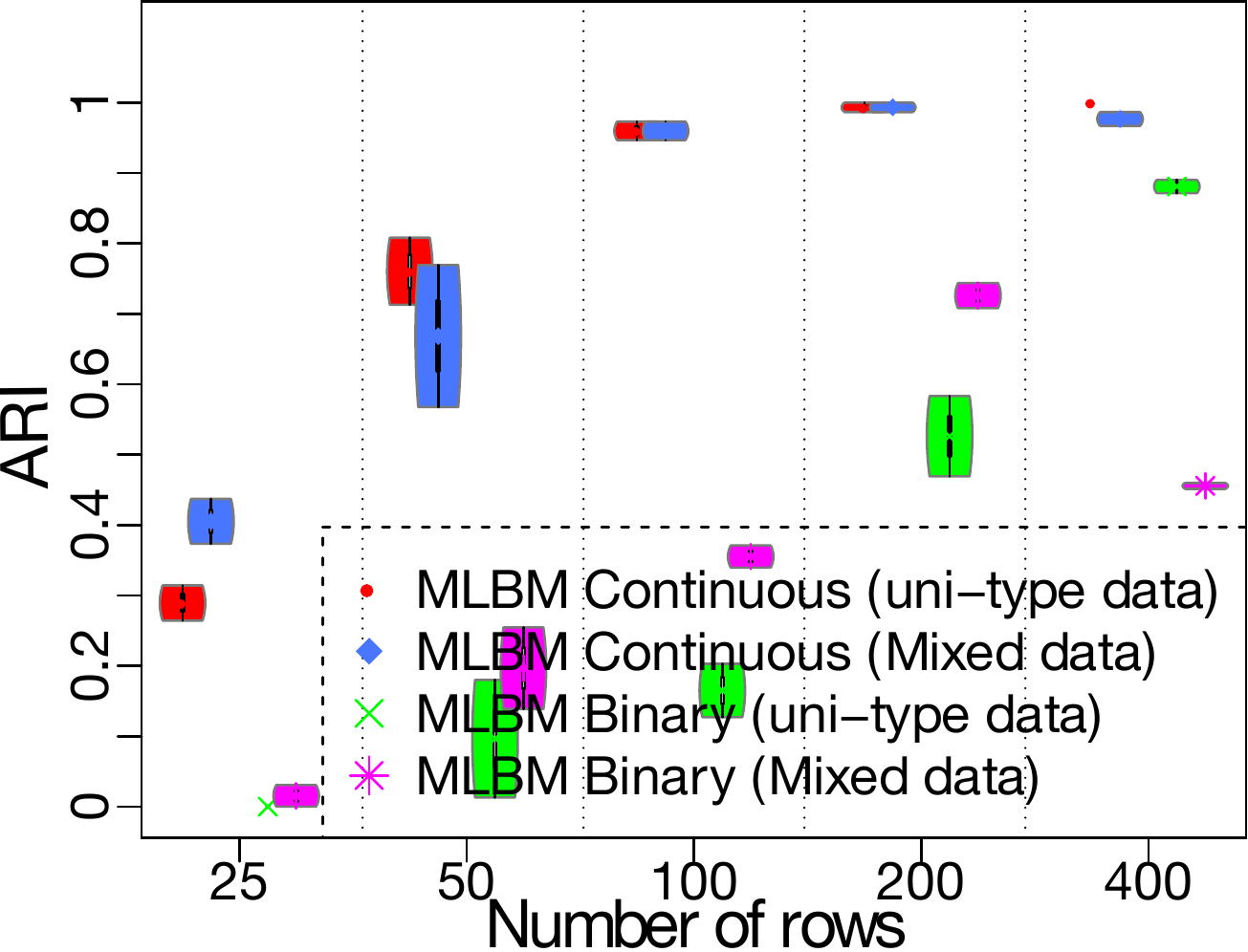}\label{p9_experiences23_high_444C}}
\caption{{\it {\small Second experiment (High confusion): ARI of rows and ARI of columns (the y-axis) in the case of continuous, binary and mixed data.}}}\label{p9_experiences23}
\end{figure}

Figures~\ref{p9_experiences21}, ~\ref{p9_experiences22} and ~\ref{p9_experiences23} show the adjusted Rand indexes of rows and columns by level of confusion and with respect to the various parameters,  in the case of continuous, binary and mixed data co-clustering. 


From the ARI plots (Figure~\ref{p9_experiences21} and Figure~\ref{p9_experiences22} in particular), the first noticeable result is that the binary part of the data is sensitive to the size of the data, to the number of co-clusters and to the level of confusion while the continuous part is generally more stable and is mostly influenced only by the number of co-clusters and the size of the data. 
\begin{itemize}
\item Influence of the number of co-clusters: given the same data size and the same level of overlap between the clusters,  we notice (Figure~\ref{p9_experiences21}) that as the number of co-clusters increases, the extraction of the true partition (both in terms of rows and column clusters) becomes harder. This effect is observed in particular in the case of binary variables as the variability of the results is greater when the number of co-clusters becomes high. However, this variability is  less drastic in the case of continuous and mixed data (see Figures \ref{p9_experiences22_medium_222R} , \ref{p9_experiences22_medium_333R}, and \ref{p9_experiences22_medium_444R} for example).  The greater the number of clusters,  the more data is required for the true partition to be found. 
 
\item Influence of the data size: the global performances of the co-clustering of uni-type data (which we have established is equivalent to the standard LBM co-clustering) confirms (as established in Section~\ref{p9_advantages}) that the co-clustering performs better as the data size increases. 
 Additionally, we notice that the continuous part of the data is always easier to co-cluster than the binary part. This is almost regardless of the data size (except in the case of large number of co-clusters: $4\times(4+4)$). The binary part on the other hand performs particularly worse for small matrices. In summary, the best partitioning of the mixed co-clusters is obtained, regardless of the number of co-clusters and the level of confusion,  with medium to large matrices. 

\item Influence of the level of confusion: the co-clustering of the mixed data performs  as expected with respect to the level of confusion. The higher the confusion, the more difficult is the extraction of the true partition of the rows, particularly in the case of small matrices (compare for example the Figures \ref{p9_experiences21_low_222R}, \ref{p9_experiences22_medium_222R}, and  \ref{p9_experiences23_high_222R}). On the contrary, even when the level of confusion is high, the quality of  the recovered co-clusters improves with the size of the data (see the evolution of the ARI values in Figure~\ref{p9_experiences23}). 

\end{itemize}

\subsubsection{The co-clustering results: rectangular matrices}
\begin{figure}[h]
\centering
\subfloat[Low confusion, $5$ columns]{\includegraphics[ width=0.31\textwidth]{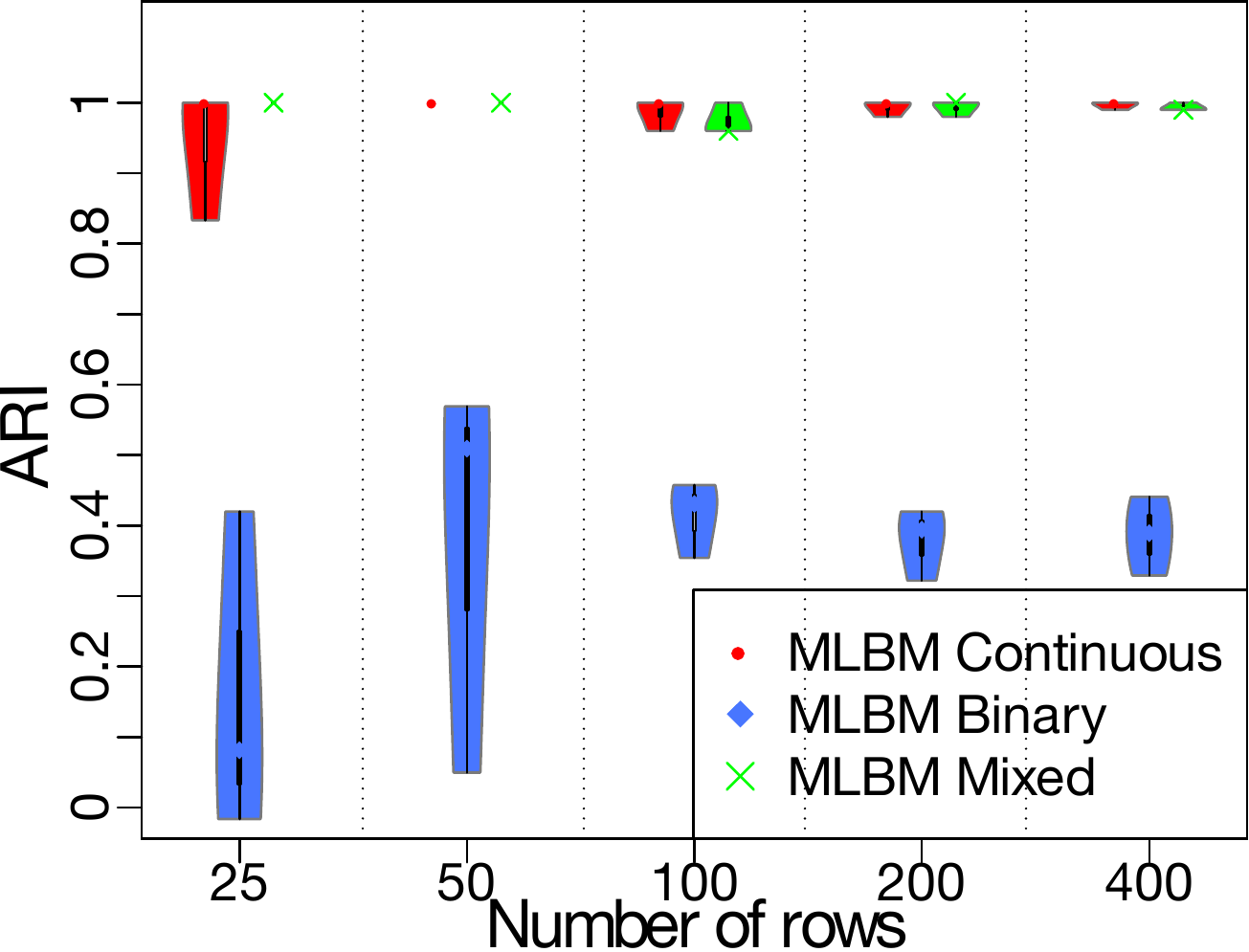}\label{p9_experiences21Rect_low_222R5}}
\quad 
\subfloat[Low confusion,  $10$ columns]{\includegraphics[width=0.31\textwidth]{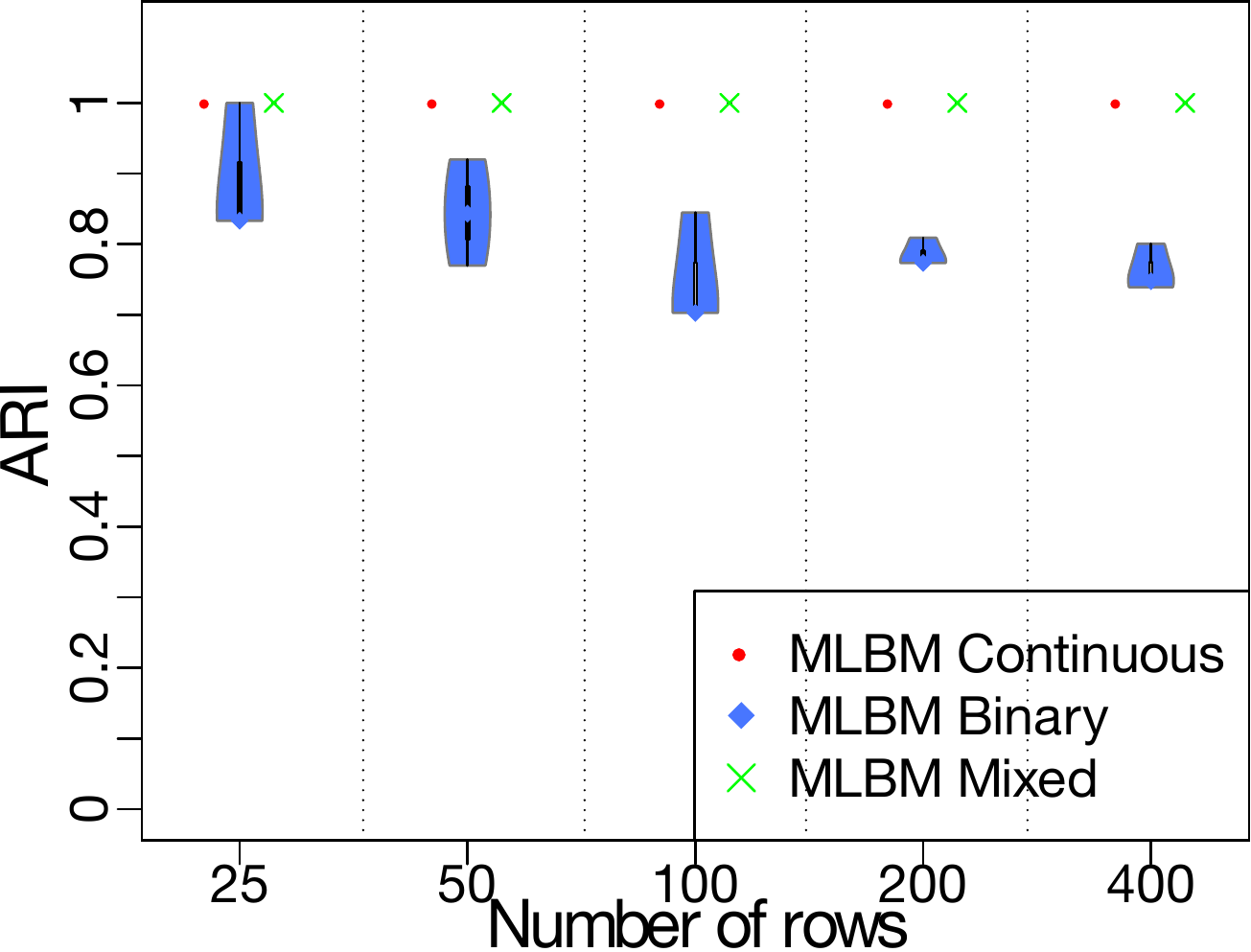}\label{p9_experiences21Rect_low_222R10}}
\quad
\subfloat[Low confusion,  $20$ columns]{\includegraphics[width=0.31\textwidth]{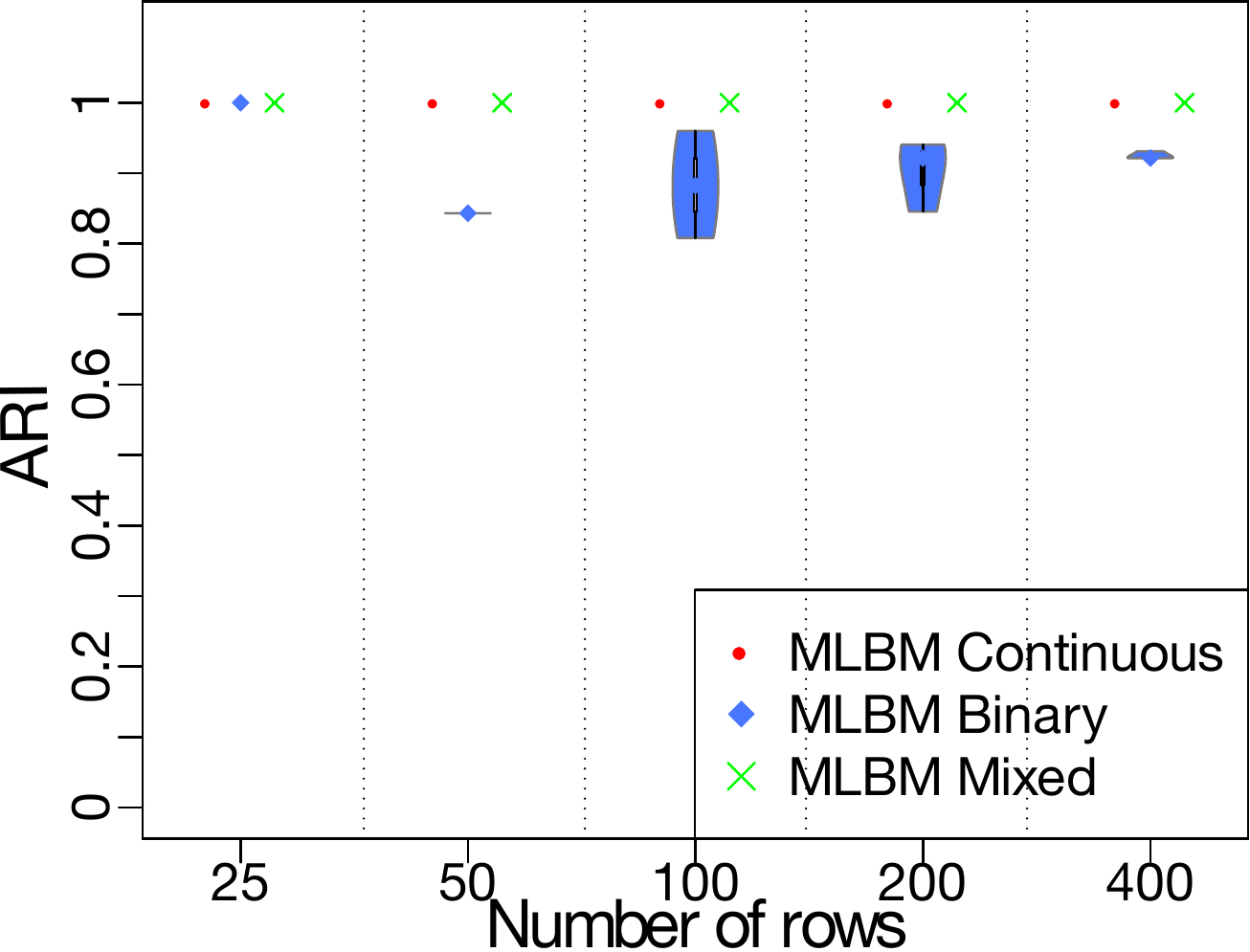}\label{p9_experiences21Rect_low_222R20}}
\quad
\subfloat[Medium confusion, $5$ columns]{\includegraphics[width=0.31\textwidth]{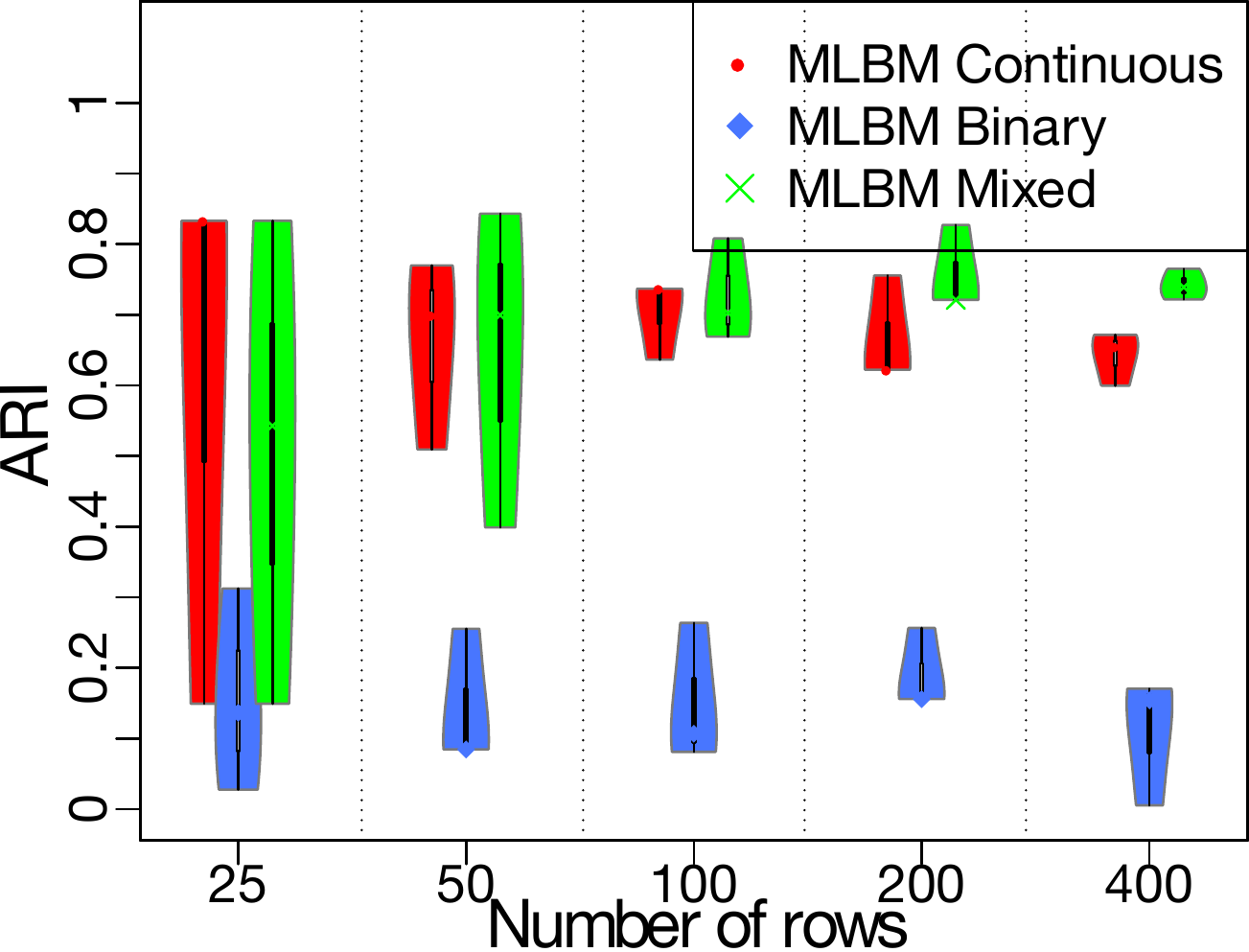}\label{p9_experiences22Rect_medium_222R5}}
\quad
\subfloat[Medium confusion, $10$ columns]{\includegraphics[width=0.31\textwidth]{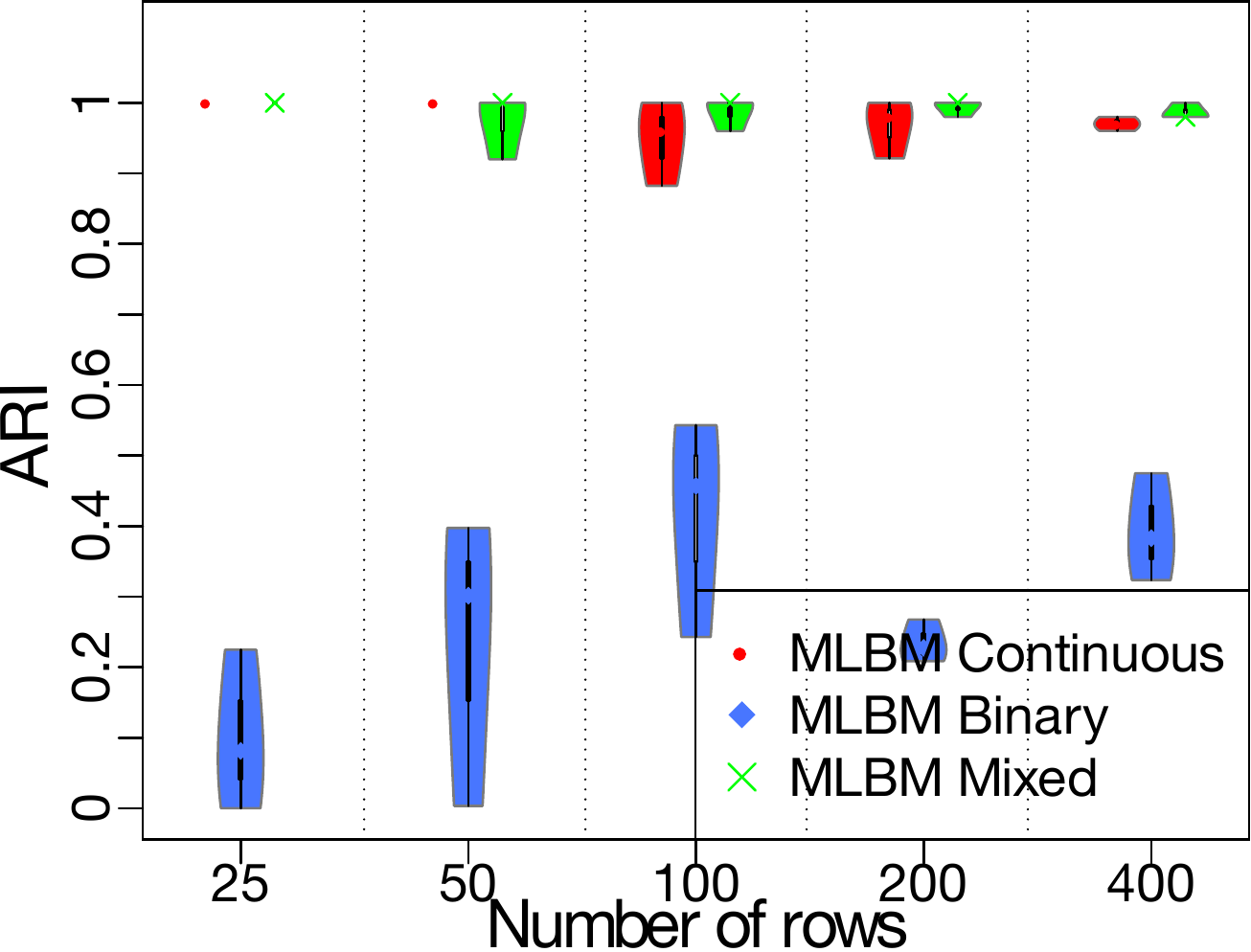}\label{p9_experiences22Rect_medium_222R10}}
\quad
\subfloat[Medium confusion, $20$ columns]{\includegraphics[width=0.31\textwidth]{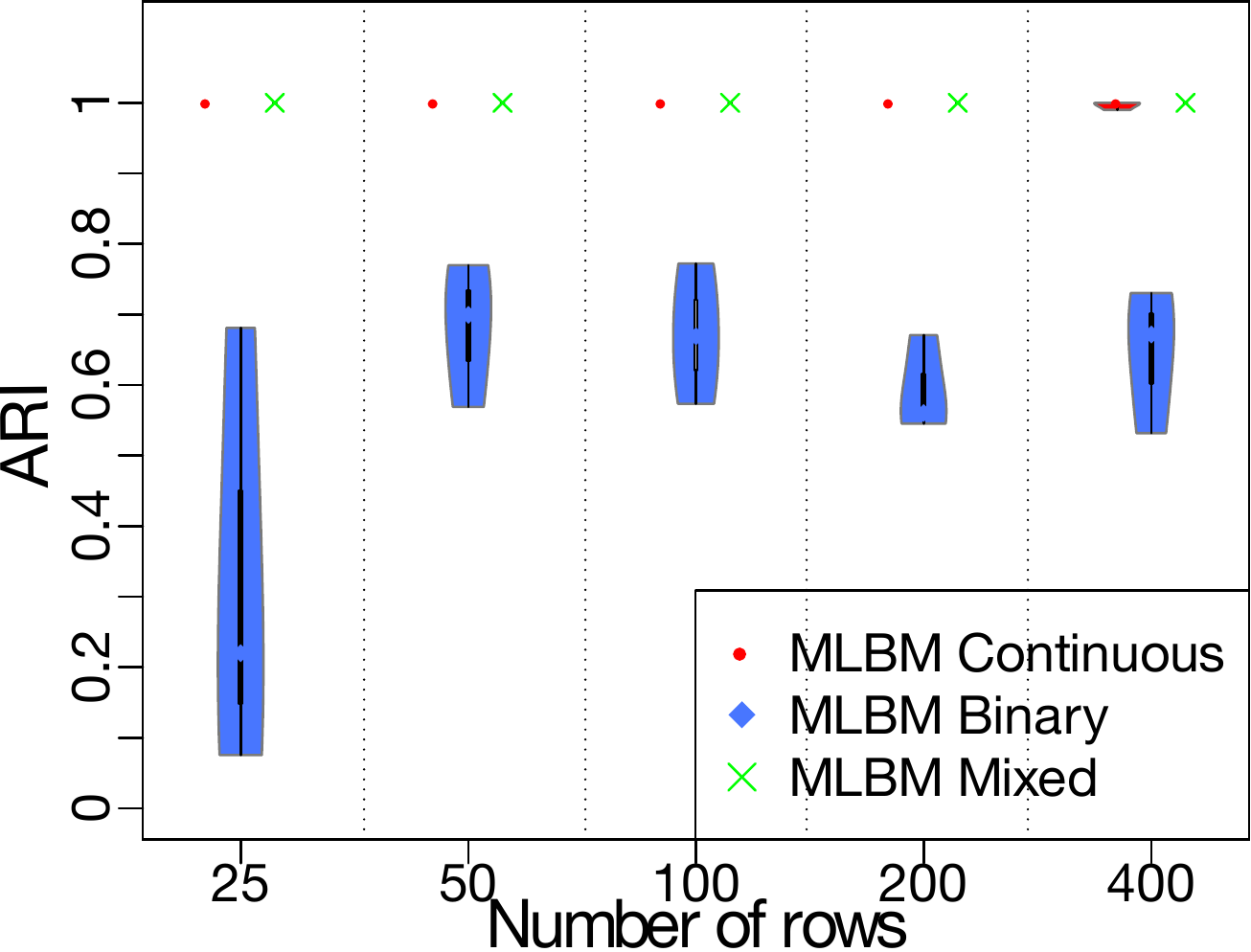}\label{p9_experiences22Rect_medium_222R20}}
\quad
\subfloat[High confusion,  $5$ columns]{\includegraphics[width=0.31\textwidth]{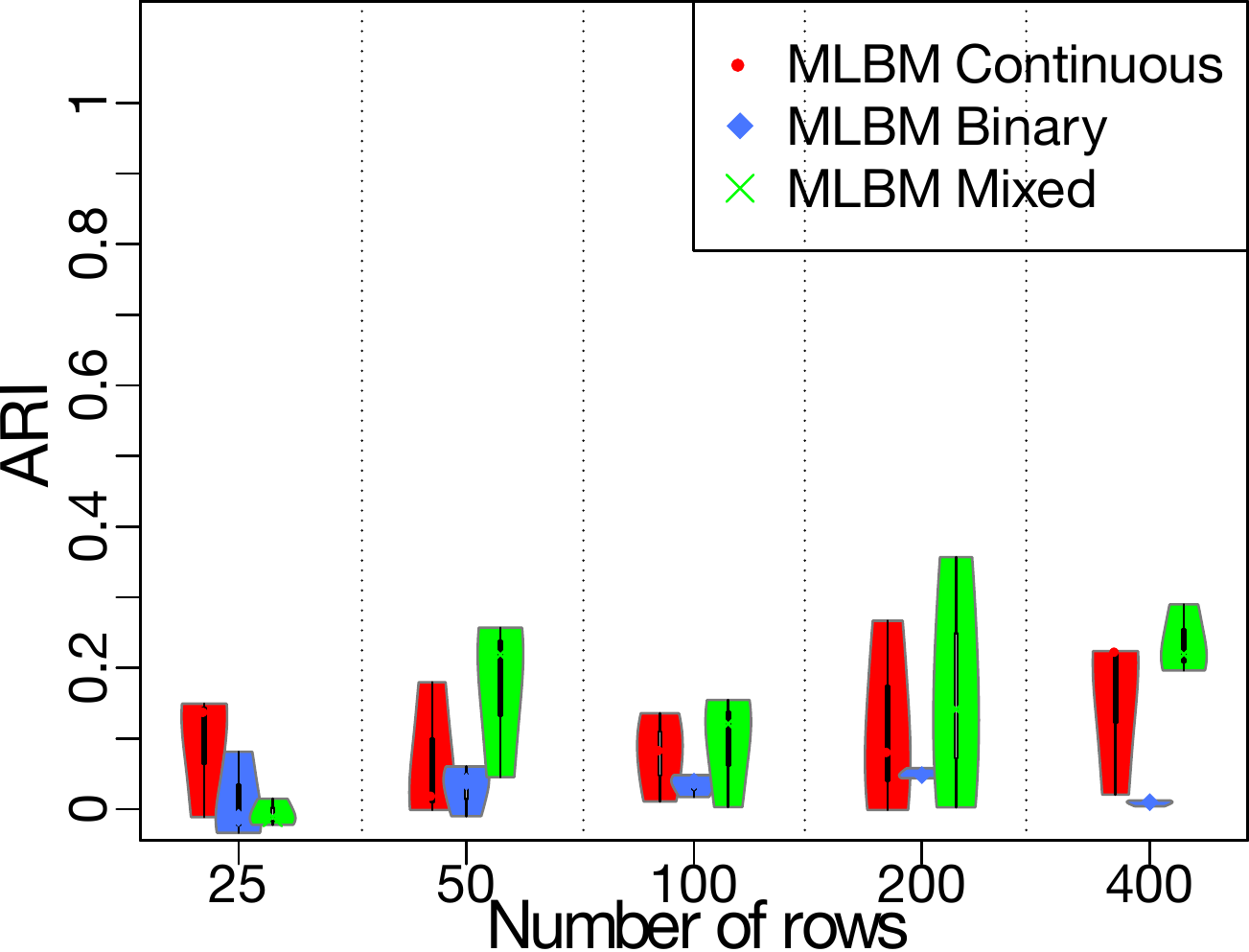}\label{p9_experiences23Rect_high_222R5}}
\quad
\subfloat[High confusion,  $10$ columns]{\includegraphics[width=0.31\textwidth]{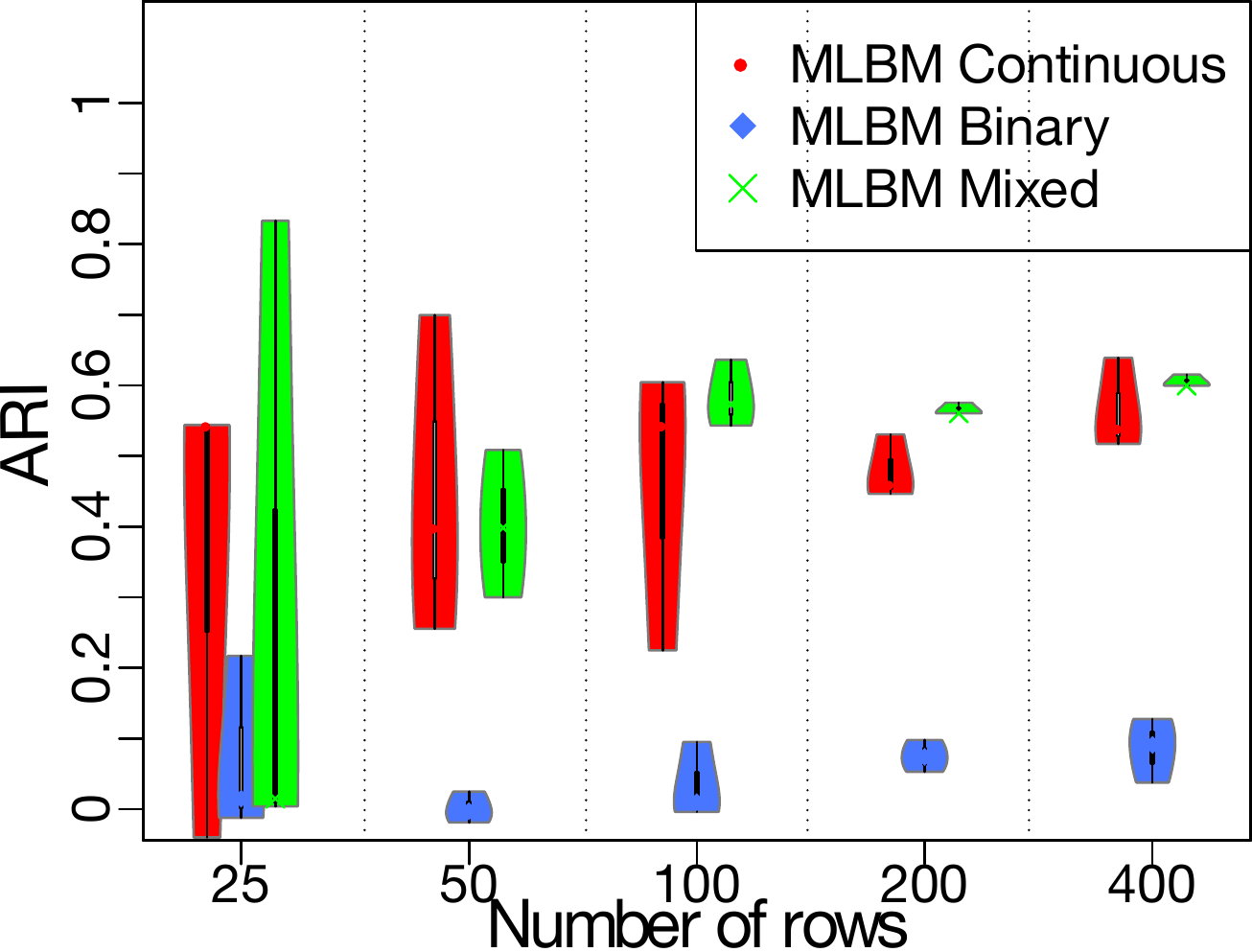}\label{p9_experiences23Rect_high_222R10}}
\quad 
\subfloat[High confusion,  $20$ columns]{\includegraphics[width=0.31\textwidth]{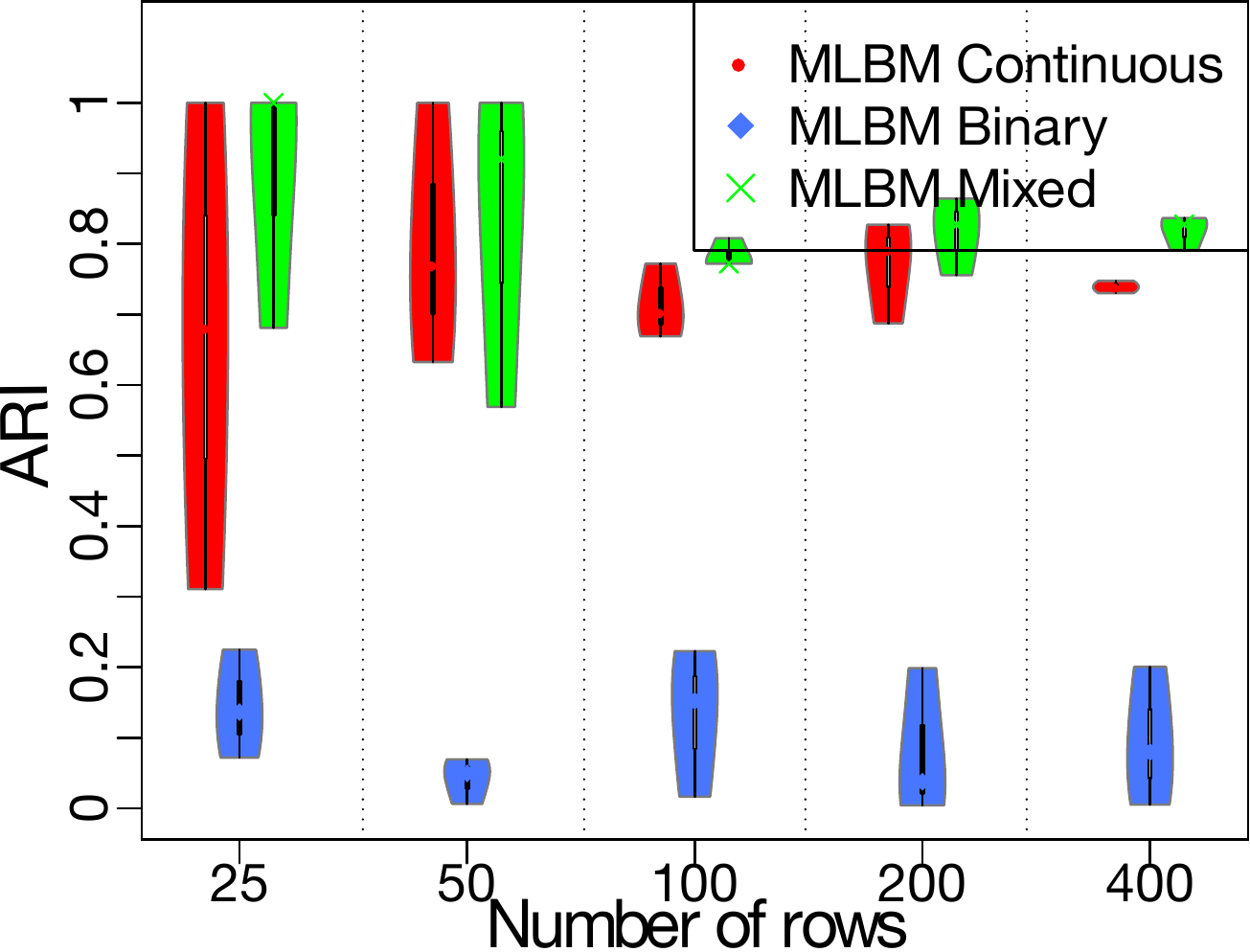}\label{p9_experiences23Rect_high_222R20}}
\caption{{\it {\small Second experiment ($2\times(2+2)$ co-clusters): ARI of rows (the y-axis) in the case of continuous, binary and mixed data.}}}\label{p9_experiences23Rect}
\end{figure}

Figure~\ref{p9_experiences23Rect} shows the adjusted Rand indexes of rows by level of confusion and with respect to the various parameters,  in the case of rectangular matrices and $2\times(2+2)$ co-clusters.  

From this experiment, we notice that even for rectangular matrices, the same conclusions are valid. In particular, the proposed approach extracts the true structure of the data in the case of low confusion. As the level of overlap between the co-clusters increases, the co-clustering of the binary part becomes less accurate both in the case of a standard LBM on uni-type data and  the case of mixed data. Finally, the bigger the data size, the more accurate is the co-clustering both using uni-type and mixed data. 
 As with the square matrices, an improvement in the ARI of columns is also noticed when using mixed data.  The same conclusions are valid for the configurations containing $3\times(3+3)$ and $ 4\times(4+4)$ co-clusters. 

\section{Discussion}\label{p9_discussion}
When applying the co-clustering algorithm on uni-type data, we noticed some optimization problems. 
 Firstly, the algorithm converges to a local optimum which corresponds, very often, to a unique cluster of rows and a unique cluster of columns. We have thus addressed the problem by forcing a minimal number of iterations (the $c$ parameter in  Algorithm~\ref{p9_LBVEM}) which considerably enhanced the quality of the optimization results.

\begin{table}[htbp]
\begin{center}
\scalebox{0.7}{
\begin{tabular}{c}
\begin{tabular}{lll}
\begin{tabular}{|l|l|l|l|}
  \hline
 $\mu$ or $\alpha$ & $J_1$ &$J_2$\\
  \hline
 $I_1$ & $p_1$ & $p_2$ \\
  \hline
  $I_2$ & $p_2$ & $p_1$ \\
  \hline
\end{tabular}& \begin{tabular}{|l|l|l|l|}
  \hline
 $\mu$ or $\alpha$  & $J_1$ &$J_2$ &$J_3$\\
  \hline
 $I_1$ & $p_1$ & $p_1$ & $p_2$\\
  \hline
  $I_2$ &$p_1$ & $p_2$ & $p_1$\\
      \hline
$I_3$ & $p_2$& $p_1$ & $p_1$ \\
  \hline
\end{tabular} &\begin{tabular}{|l|l|l|l|l|}
  \hline
$\mu$ or $ \alpha$  & $J_1$ &$J_2$ &$J_3$&$J_4$\\
  \hline
 $I_1$ & $p_1$&  $p_1$ &  $p_1$& $p_2$\\
  \hline
  $I_2$ & $p_1$ &  $p_1$ &  $p_2$& $p_1$\\
      \hline
$I_3$ &  $p_1$ &  $p_2$ &  $p_1$ & $p_1$\\
      \hline
$I_4$ &  $p_2$ &  $p_1$ &  $p_1$ & $p_1$\\
  \hline
\end{tabular}
\end{tabular}\\
\end{tabular}
}\caption{{\small{{\it The true specification of the co-clusters in a symmetric configuration with $2\times(2+2)$, $3\times(3+3)$ and $4\times(4+4)$ co-clusters.}}}}\label{p9_configurations3}
 \end{center}
\end{table}

However, the algorithms (both our approach and the blockcluster package) do not perform the same way when the marginal parameters are equal per cluster or when they are different.  To study this effect, we have considered a second configuration (call it the {\it symmetric} case)  where the marginal parameters are equal.  Table~\ref{p9_configurations3} shows an example of the parameter specification of such configurations.  In the {\it symmetric} configuration, where the marginal parameters are equal, the problem of cluster separability becomes intrinsically difficult (especially for square matrices) and the optimization algorithm tends to have trouble getting out of the zone of the local optimum corresponding to one single cluster of rows and one single cluster of columns, in which it falls since the very first iterations. To solve this problem, we required the algorithm to start with small steps when computing the assignments  to the clusters ($s$, $tc$ and $td$) without letting the criterion fully stabilize, then after few first steps in this initial phase, we iterate until criterion stabilization. This strategy provides better solutions in the case of binary data but results in no notable improvement in some continuous cases. 
 As mentioned earlier, because of this focus on obtaining high quality results, our implementation takes at least ten times longer than
the blockcluster package but provides more accurate row and column partitions and more accurate parameter estimation. Table~\ref{p9_computationTime} shows a comparative example of the means computation time for the rectangular matrix containing $100$ rows  and $2\times(2+2)$ co-clusters.

\begin{table}[htbp]
\begin{center}
\scalebox{0.8}{
\begin{tabular}{c}
\begin{tabular}{|l|c|c|c|c|}
\hline
Number of columns & $\rightarrow$ & 5 columns & 10 columns & 20 columns\\
 \hline
\begin{tabular}{l}
Level of overlap\\
 \hline
Low confusion\\
\\
\hline
Medium confusion\\
\\
\hline
High confusion\\
\\
 \end{tabular}  & \begin{tabular}{c}
 measure$\backslash$method\\
 \hline
mean\\
sd\\
\hline
mean\\
sd \\
\hline
mean\\
sd\\
 \end{tabular} & \begin{tabular}{c|c}
 MLBM & BC\\
 \hline
 $2.97$ &$0.01$\\
$0.11$ & $0.005$ \\
\hline
$8.01$ &$0.01$\\
$1.2$ & $0.01$ \\
\hline
$15$ &$0.04$\\
$12$ & $0.01$ \\

 \end{tabular} &\begin{tabular}{c|c}
 MLBM & BC\\
 \hline
$4.3$ &$0.016$\\
$0.4$ & $0.005$ \\
\hline
$5.5$ &$0.01$\\
$0.2$ & $0$ \\
\hline
$16.4$ &$0.01$\\
$6.4$ & $0.01$ \\

 \end{tabular} & \begin{tabular}{c|c}
 MLBM & BC\\
 \hline
 $8.31$ &$0.03$\\
$0.6$ & $0.005$ \\
\hline
$8.4$ &$0.01$\\
$0.3$ & $0$ \\
\hline
$25.9$ &$0.01$\\
$5.2$ & $0.01$ \\

 \end{tabular}\\
\hline
\end{tabular}
\end{tabular}
}\caption{{\small{{\it Example of the computation time (in seconds).}}}}\label{p9_computationTime}
 \end{center}
\end{table}

\section{Conclusion and future work}\label{p9_conclusion}
In this article, we have proposed an extension of the latent block models to the co-clustering of mixed type data.  The experiments show the capability of the approach to estimate the true model parameters, extract the true distributions from simulated data, and provide better quality results when the complete data set is used rather than separately co-clustering the continuous or binary parts. The proposed approach comes as a natural extension of the LBM based co-clustering and performs a co-clustering of mixed data in the same way that a standard LBM based co-clustering applies to uni-type data. 

 On the course of our experiments, we have noticed that for the data sets with equal marginal parameters, both our algorithm and the state of the art algorithm implemented in the package blockcluster tend to fall in a local optimum. This is a limitation to the latent block based methods for co-clustering, mainly in the context of an exploratory analysis where the true underlying distributions are unknown.
 
  In our future works, we aim to extend the approach to the case of categorical data and beyond binary data and to study the option of BIC based regularization to automatically infer the number of clusters of rows and the number of clusters of columns.

\bibliographystyle{apalike}
\bibliography{biblio}

\begin{thebibliography}{}

\bibitem[Bhatia et~al., 2014]{bhatia2014}
Bhatia, P., Iovleff, S., and Govaert, G. (2014).
\newblock blockcluster: An r package for model based co-clustering.
\newblock working paper or preprint : \url{https://hal.inria.fr/hal-01093554}.

\bibitem[Bock, 1979]{bock1979}
Bock, H. (1979).
\newblock Simultaneous clustering of objects and variables.
\newblock In {\em E. Diday (ed) Analyse des données et Informatique}, page
  187–203. INRIA.

\bibitem[Brault and Lomet, 2015]{brault2015b}
Brault, V. and Lomet, A. (2015).
\newblock Revue des méthodes pour la classification jointe des lignes et des
  colonnes d’un tableau.
\newblock {\em Journal de la Société Française de Statistique},
  156(3):27--51.

\bibitem[Cheng and Church, 2000]{church2000}
Cheng, Y. and Church, G.~M. (2000).
\newblock Biclustering of expression data.
\newblock In {\em Proceedings of the International Conference on Intelligent
  Systems for Molecular Biology}, volume~8, pages 93--103. AAAI Press.

\bibitem[Dhillon et~al., 2003]{dhillon2003}
Dhillon, I.~S., Mallela, S., and Modha, D.~S. (2003).
\newblock Information-theoretic co-clustering.
\newblock In {\em Proceedings of the ninth international conference on
  Knowledge discovery and data mining}, pages 89--98. ACM Press.

\bibitem[Good, 1965]{good1965}
Good, I.~J. (1965).
\newblock Categorization of classification.
\newblock In {\em Mathematics and Computer Science in Biology and Medicine},
  pages 115--125. Her Majesty's Stationery Office, London.

\bibitem[Govaert and Nadif, 2003]{govaert2003}
Govaert, G. and Nadif, M. (2003).
\newblock Clustering with block mixture models.
\newblock {\em Pattern Recognition}, 36(2):463--473.

\bibitem[Govaert and Nadif, 2008]{govaert2008}
Govaert, G. and Nadif, M. (2008).
\newblock Block clustering with {B}ernoulli mixture models : Comparison of
  different approaches.
\newblock {\em Computational Statistics and Data Analysis}, 52(6):3233--3245.

\bibitem[Govaert and Nadif, 2013]{govaert2013}
Govaert, G. and Nadif, M. (2013).
\newblock {\em Co-Clustering}.
\newblock ISTE Ltd and John Wiley \& Sons Inc.

\bibitem[Hartigan, 1975]{hartigan1975}
Hartigan, J.~A. (1975).
\newblock {\em Clustering Algorithms}.
\newblock John Wiley \& Sons, Inc., New York, NY, USA.

\bibitem[Hintze and Nelson, 1998]{violin_plot}
Hintze, J.~L. and Nelson, R.~D. (1998).
\newblock Violin plots: A box plot-density trace synergism.
\newblock {\em The American Statistician}, 52(2):181--184.

\bibitem[Hubert and Arabie, 1985]{Hubert1985}
Hubert, L. and Arabie, P. (1985).
\newblock Comparing partitions.
\newblock {\em Journal of Classification}, 2(1):193--218.

\bibitem[Mariadassou and Matias, 2015]{Matias2015}
Mariadassou, M. and Matias, C. (2015).
\newblock {Convergence of the groups posterior distribution in latent or
  stochastic block models}.
\newblock {\em {Bernoulli}}, 21(1):537--573.

\bibitem[McParland and Gormley, 2016]{damien2015}
McParland, D. and Gormley, I.~C. (2016).
\newblock Model based clustering for mixed data: clustmd.
\newblock {\em Advances in Data Analysis and Classification. Springer},
  10(2):155--169.

\end{thebibliography}

\end{document}